\documentclass[letterpaper]{article} 
\usepackage[preprint]{aaai2027} 
\usepackage[hyphens]{url}  
\usepackage{graphicx} 
\usepackage{natbib}  
\usepackage{caption} 
\usepackage{amsmath}
\usepackage{amssymb}
\usepackage{booktabs}
\usepackage{array}
\usepackage{multirow}
\usepackage{float}

\title{Beyond Transformers: Linear Attention Policy for Open-Vocabulary Object Goal Navigation}

\author{
Jiahong Zhang\textsuperscript{\rm 1,\rm 2}\equalcontrib,
Yifan Lin\textsuperscript{\rm 1,\rm 3}\equalcontrib,
Yandong Zhang\textsuperscript{\rm 4},
Sijun Shen\textsuperscript{\rm 1,\rm 3},
Kexin Wang\textsuperscript{\rm 1,\rm 2},
Yuqi Pan\textsuperscript{\rm 1,\rm 2},
Hongjuan Pei\textsuperscript{\rm 5},
Wei Wang\textsuperscript{\rm 4},
Guoqi Li\textsuperscript{\rm 1}\corresponding
}

\affiliations{
\textsuperscript{\rm 1}Institute of Automation, Chinese Academy of Sciences, Beijing 100045, China\\
\textsuperscript{\rm 2}School of Artificial Intelligence, University of Chinese Academy of Sciences, Beijing 100049, China\\
\textsuperscript{\rm 3}State Key Laboratory of Media Convergence and Communication, Communication University of China, Beijing 100024, China\\
\textsuperscript{\rm 4}Beihang University, Beijing 100191, China\\
\textsuperscript{\rm 5}University of Chinese Academy of Sciences, No.19A Yuquan Road, Beijing 100049, China\\
zhangjiahong2023@ia.ac.cn,
lyfgin1002@gmail.com,
yandongzhang@buaa.edu.cn,
sijun.shen@mails.cuc.edu.cn,\\
wangkexin2021@ia.ac.cn,
panyuqi2024@ia.ac.cn,
peihongjuan@ucas.ac.cn,
w.wang@buaa.edu.cn,
guoqi.li@ia.ac.cn
}

\begin{document}
\maketitle

\begin{abstract}
Open-Vocabulary Object Goal Navigation (OVON) requires agents to operate under partial observability, making effective internal state updates critical for navigation performance. This update is implemented by the policy network, where recent approaches adopt Transformer-based backbones with self-attention over a context window to integrate temporal information. However, our controlled experiments show that performance does not scale with context length under Transformer-based policies, questioning the suitability of self-attention for state integration in navigation. To this end, we propose Linear Attention-based Navigation (LANav), which adopts linear attention(LA) as the policy backbone to maintain a structured state update rather than self-attention over the context window. Across multiple LA variants evaluated under identical settings, LANav consistently outperforms Transformer-based baselines. Performance improves as state update mechanisms become more structured and regulated, highlighting the importance of state update design.
To improve state update effectiveness, we introduce Weighted State-Expansion Linear Attention (WSLA), which expands each attention head’s state into multiple sub-states
and uses learnable weighted readout to aggregate expanded sub-states. Equipped with WSLA, LANav achieves 36.4\% average success rate (SR) on HM3D-OVON, outperforming Transformer-based counterparts by 6.3 percentage points in macro-averaged SR, while maintaining computational efficiency. Distance-stratified results show larger gains in long-distance episodes, while HSSD transfer and fine-tuning demonstrate robustness across scene distributions. Real-world deployment on a Unitree Go2 further achieves an 82\% success rate over 50 trials, supporting the practical feasibility and sim-to-real transfer of LANav.
\end{abstract}

\section{Introduction}

Object Goal Navigation (ObjectNav) serves as a representative benchmark in embodied intelligence, where an agent must explore an unseen environment and locate an object belonging to a specified target category~\cite{objectnav}. Such exploration under partial observability requires the policy to maintain an internal state that summarizes past observations.
Navigation agents therefore commonly adopt recurrent neural networks (RNNs) as the policy backbone, where temporal information is encoded into compact hidden states for action execution~\cite{ye2021auxiliary,pirlnav,xgx}. However, recurrent architectures such as Long Short-Term Memory (LSTM)~\cite{hochreiter1997long} and Gated Recurrent Units (GRU)~\cite{cho2014learning} struggle to preserve dependencies over extended trajectories~\cite{fang2019scene,fortunato2019generalization}.
To address these limitations, Transformer-based policies introduce explicit self-attention~\cite{vaswani2017attention} over a fixed-length context window, enabling direct interaction among historical observations~\cite{fukushima2022object,li2023transformer}. Despite this design, their performance remains limited on the challenging Open-Vocabulary Object Goal Navigation (OVON)~\cite{ovon}.

Open-Vocabulary settings introduce greater semantic ambiguity and generalization difficulty, placing higher demands on effective state updating of agents. Our controlled study of Transformer-based policies on OVON reveals that performance does not improve as the context length increases. This finding motivates us to explore alternative state update mechanisms for improving performance.

This work investigates LA that replaces self-attention in Transformers with kernelized or state-space formulations, enabling structured state updates over the context window~\cite{katharopoulos2020transformers,gla,chou2024metala}.
LA was originally proposed to reduce the computational complexity of standard Transformers, and has demonstrated strong scalability in long-sequence tasks such as language modeling~\cite{deltanet,sun2023retentive,yanggated} and decision modeling~\cite{david2022decision,dai2024mamba,sun2025memory}.
This paper examines LA from a different perspective. We show that its state update mechanism is better suited for integration of historical information in navigation. Our empirical study reveals three key findings:
(i) LA-based policies consistently outperform both RNN and Transformer baselines.
(ii) LA benefits from larger training context lengths, whereas Transformer-based policies do not gain from longer training contexts.
(iii) Among different LA variants, models with more effective state update mechanisms achieve stronger performance.

Based on these findings, we introduce Linear Attention-based Navigation (LANav), which adopts LA as the policy backbone. We further design a Weighted State-Expansion Linear Attention (WSLA) to improve state update effectiveness in navigation.
WSLA is inspired by prior studies on state expansion and multi-state modeling in linear-time sequence models~\cite{liu2025scaling,wang2025mmdend,pan2025spikingbrain}. It expands each attention head into multiple sub-states and introduces learnable weighting to regulate their readout contributions, thereby enhancing state expressivity in navigation.
Built upon WSLA, LANav achieves superior performance on
Habitat-Matterport3D Open-Vocabulary ObjectNav (HM3D-OVON)~\cite{ovon}
while maintaining favorable computational efficiency and scalability.

The contributions of this work are summarized as follows:

(1) We introduce LA as the policy backbone for OVON, termed \textbf{LANav}. Through controlled comparisons under matched architectures and training settings, we show that LA backbones consistently outperform both RNN and Transformer baselines. These findings establish LA as an effective mechanism for sequential navigation decision-making and reveal a promising design direction for future embodied policy.

(2) We provide a comprehensive comparison within the LA family, including Linear Transformer~\cite{schlag2021linear}, DeltaNet~\cite{deltanet}, and Gated DeltaNet~\cite{yanggated}. Our study reveals that progressively more structured and regulated state update rules lead to consistent performance improvements in navigation.

(3) We propose \textbf{WSLA}, which expands the state of LA into multiple sub-states and introduces learnable weighting to regulate sub-state contributions during readout. Built upon WSLA, LANav achieves an average SR of \textbf{36.4\%} on HM3D-OVON, outperforming the strongest Transformer-based baseline (30.1\%) under comparable settings.

\section{Related Work}

\subsection{Policy Backbones for Open-Vocabulary Object Goal Navigation}

OVON extends ObjectNav to language-specified categories that may be unseen during training or described
by novel names and synonyms~\cite{ovon}, making effective policy state updates
under partial observability critical for long-horizon navigation. Existing
OVON-related methods include map-based and mapless learned
policies. Map-based methods maintain explicit spatial or semantic
representations for exploration and planning
~\cite{chaplot2020semanticexploration}. Recent open-vocabulary
systems further combine vision-language models with detectors,
segmenters, or structured exploration modules
~\cite{zemskova2025ovsegdtsegmentingtransformeropenvocabulary,
ziliotto2025tango,zhu2025move}. However, these systems often differ
in sensors, external modules, supervision, and training protocols,
making controlled backbone comparison difficult.

Mapless policies predict actions from egocentric observation sequences and rely
on the policy network to maintain an implicit state. Recurrent backbones such as GRU and LSTM are widely used in navigation
policies~\cite{mirowski2017learning,wijmansdd,pirlnav,mcfc,
ramrakhya2022habitat,yadav2023offline}, but struggle with long-range
dependencies~\cite{fang2019scene,fortunato2019generalization}.
Transformer-based policies use self-attention over a context window and have
been explored in visual navigation and control~\cite{ye2021auxiliary,
chen2021decisiontransformer,li2023transformer,lawson2023control,
wang2024navformer,zeng2025poliformer}. In OVON, they outperform recurrent
baselines~\cite{ovon,zemskova2025ovsegdtsegmentingtransformeropenvocabulary},
yet how temporal backbones should integrate extended histories remains
underexplored.

Our work addresses this gap by isolating the policy backbone as the key variable.
By fixing the visual and language encoders, navigation data, and training pipeline,
we study how different temporal backbones update policy states under partial
observability. This enables a controlled comparison of RNN, Transformer, and
linear-attention backbones for history integration in OVON.

\subsection{Linear Attention for Decision-Making}

LA was introduced to reduce the quadratic cost of Transformer
self-attention by replacing softmax attention with kernelized or gated
formulations~\cite{schlag2021linear,katharopoulos2020transformers,gla}.
Its recurrent form updates an internal state incrementally, reducing the
attention complexity from $O(l^2)$ to $O(l)$ with respect to the context length
$l$. This property has made LA effective for long-sequence
modeling~\cite{sun2023retentive,pan2025scaling}.

Linear-time sequence models have also been studied in decision-making and
reinforcement learning. Decision Transformer~\cite{chen2021decisiontransformer}
shows the benefit of sequence modeling for control, while Decision
S4~\cite{david2022decision} and Decision
Mamba~\cite{ota2024decisionmamba} demonstrate that state-space or selective-scan
updates can improve efficiency while retaining competitive performance.

Different from prior work that mainly emphasizes efficiency, we study linear
attention as a state-update mechanism for embodied navigation. RNNs compress
history into a single hidden state, while Transformers perform self-attention
over a fixed context window. LA offers a third design choice by
maintaining structured state updates derived from attention factorization, which
may better support history integration in partially observable navigation.

\begin{figure*}[t]
\centering
\includegraphics[width=1\textwidth]{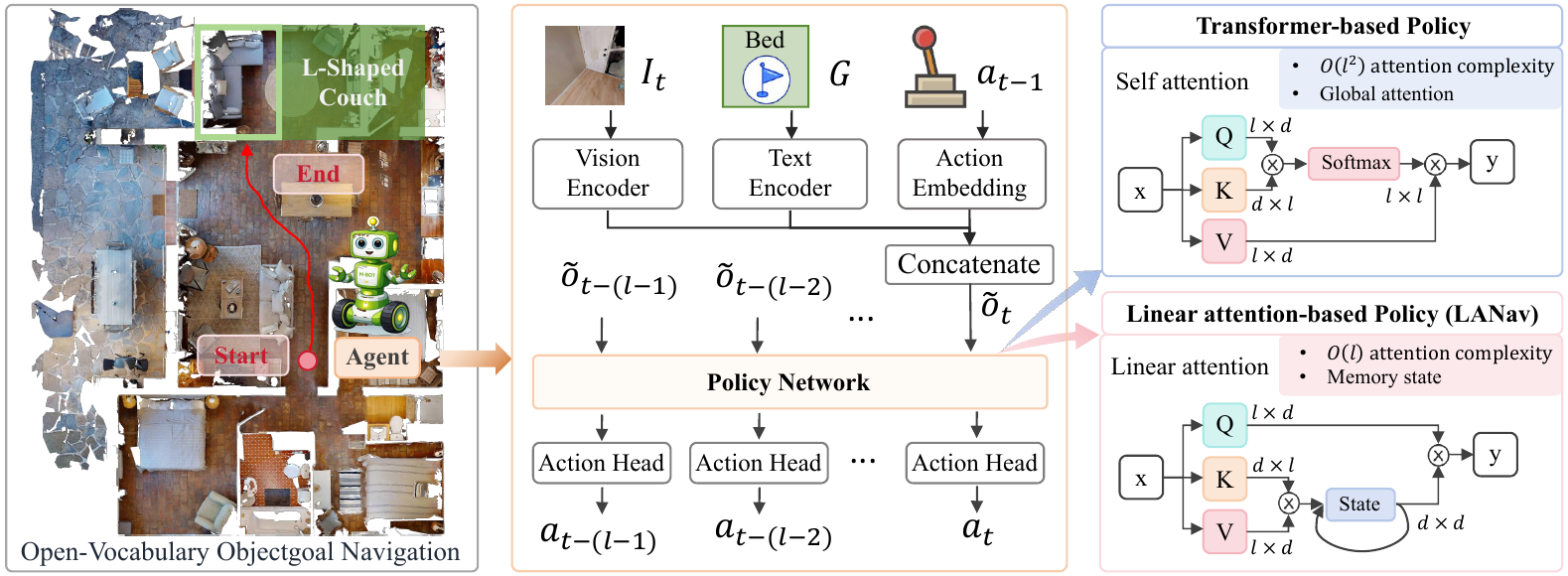}
\caption{\textbf{Overview of LANav.}
\textbf{Left:} Open-Vocabulary Object Goal Navigation task.
\textbf{Middle:} At step $t$, the RGB observation $I_t$, goal description $G$, and previous action $a_{t-1}$ are encoded by a vision encoder, a text encoder, and an action embedding module, respectively. The resulting features form a per-step embedding $\tilde{o}_t$. A context window consisting of the most recent embeddings $\{\tilde{o}_{t-(l-1)}, \dots, \tilde{o}_t\}$ is processed by the policy backbone to produce the current representation and predict the action $a_t$.
\textbf{Right:} Comparison of policy backbones. Transformer-based policies perform self-attention over the full context, whereas LA-based policies maintain an internal state that is incrementally updated.}
\label{fig:fig1}

\end{figure*}

\section{Method}

\subsection{Problem Formulation}

We study the OVON task, where a mobile agent navigates in an indoor environment to find an object specified by language. Unlike closed-set ObjectNav, the target category may be unseen during training or described by a novel name or a semantically related synonym. The agent must therefore generalize beyond a fixed goal vocabulary. Partial observability further increases the difficulty. The agent only receives egocentric observations, and goal-relevant cues may appear only briefly during exploration. Meanwhile, many observations can be redundant or unrelated to the target. A navigation policy must retain useful semantic evidence while limiting the influence of irrelevant history.


At each decision step $t$, the agent receives an egocentric RGB observation
$I_t$, a language-specified goal category $G$, and the previous action
$a_{t-1}$.
Following the policy setting of DAgRL~\cite{ovon}, no explicit pose or
egomotion observation is provided to the policy. The per-step policy input is defined as:
\begin{equation}
x_t=(I_t,G,a_{t-1}),
\label{eq:policy_input}
\end{equation}

The action space $\mathcal{A}$ is discrete and consists of move-forward,
turn-left, turn-right, look-up, look-down, and stop. At each step, the policy
selects an action $a_t \in \mathcal{A}$. The episode terminates when stop is
issued or the maximum step limit is reached, and is considered successful if
the stop action is issued within a predefined distance threshold of any object
instance from the goal category.

OVON can be formulated as a partially observable Markov
decision process (POMDP). Since the complete scene layout
and object locations are not fully observable, the agent cannot
reliably select actions from the current observation alone.
Instead, the action distribution is predicted from all inputs
available up to step $t$, including the current observation, the
language-specified goal, and previous interactions:
\begin{equation}
a_t \sim
F_\theta
\left(
\cdot \mid I_{1:t}, G, a_{0:t-1}
\right)
=
F_\theta
\left(
\cdot \mid x_{\leq t}
\right).
\end{equation}
Here, $F_\theta$ denotes the navigation policy, with $\theta$ representing its learnable parameters. The shorthand $x_{\leq t}$ collects the policy inputs available up to step $t$, including the current input $x_t$ and the historical inputs $x_{<t}$. To effectively model such sequential policy inputs, we introduce LANav, which uses linear attention to efficiently aggregate historical information and update the policy representation over time.

\subsection{Overview of LANav}

Figure~\ref{fig:fig1} illustrates the LANav architecture, which follows the
DAgRL policy design~\cite{ovon} by separating visual encoding, goal encoding,
previous-action embedding, temporal information integration, and action
prediction.

At each step $t$, the RGB observation $I_t$ is processed by
a frozen SigLIP$_{\text{RGB}}$ encoder
\cite{zhai2023sigmoid}, yielding
$v_t=f_{\mathrm{vis}}(I_t)\in\mathbb{R}^{768}$. The goal
description $G$ is encoded by a frozen
SigLIP$_{\text{text}}$ encoder as
$g=f_{\mathrm{text}}(G)\in\mathbb{R}^{768}$. The goal
embedding is computed once and remains fixed throughout the
episode. To provide behavioral context, the previous action
$a_{t-1}$ is mapped to a $32$-dimensional vector by a learnable
embedding layer, $e_t^a=f_{\mathrm{act}}(a_{t-1})\in\mathbb{R}^{32}$.

The visual, goal, and previous-action embeddings are
concatenated and linearly projected to the backbone hidden
dimension $d$:
\begin{equation}
\tilde{o}_t
=
\mathbf{W}_o [v_t; g; e_t^{a}]
+
\mathbf{b}_o,
\end{equation}
where
$\mathbf{W}_o\in\mathbb{R}^{d\times1568}$ and
$\mathbf{b}_o\in\mathbb{R}^{d}$ are learnable parameters.
We set the backbone hidden dimension to $d=512$.

Let $l$ denote the maximum context length and
$L_t=\min(l,t)$. The policy network receives up
to the $L_t$ most recent projected embeddings:
\begin{equation}
\label{eq:zt}
\mathbf{Z}_t
=
[\tilde{o}_{t-L_t+1},\ldots,\tilde{o}_t]
\in
\mathbb{R}^{L_t\times d}.
\end{equation}
This definition also covers the beginning of an episode,
when fewer than $l$ observations are available.

The resulting sequence is processed by a four-layer
decoder-only Transformer or LA backbone
$\Phi(\cdot)$:
\begin{equation}
\label{eq:lanav_state}
\mathbf{y}_t
=
\bigl[\Phi(\mathbf{Z}_t)\bigr]_{L_t}
\in\mathbb{R}^{d},
\end{equation}
where $\mathbf{y}_t$ is the backbone output corresponding to the most
recent time step. 
Finally, the action head projects $\mathbf{y}_t$ to a categorical
distribution over the discrete action space:
\begin{equation}
\pi_\theta
\left(
\cdot \mid I_{1:t},G,a_{0:t-1}
\right)
=
\operatorname{Softmax}
\left(
\mathbf{W}_a \mathbf{y}_t+\mathbf{b}_a
\right),
\end{equation}
where
$\mathbf{W}_a\in
\mathbb{R}^{|\mathcal{A}|\times d}$ and
$\mathbf{b}_a\in\mathbb{R}^{|\mathcal{A}|}$ are learnable
parameters of the action head.

\subsection{Policy Backbones}
\label{subsec:Policy Backbones}

The policy backbone $\Phi(\cdot)$ in Eq.~(5) integrates sequential policy inputs and produces the policy representation $\mathbf{y}_t$ at the current navigation step. We compare a standard causal Transformer with linear-attention backbones and examine how they differ in maintaining and accessing historical information.

\subsubsection{Transformer backbone.}
We use a lightweight decoder-only Transformer with causal self-attention as
the baseline. Let $\mathbf{z}_\tau$ denote the token representation at step
$\tau$ in the current backbone layer; at the first layer,
$\mathbf{z}_\tau=\tilde{\mathbf{o}}_\tau$. Each attention head $h$ projects
the token into query, key, and value vectors:
\begin{equation}
\mathbf{q}_\tau^{(h)}
=
\mathbf{W}_{q}^{(h)}\mathbf{z}_\tau,\quad
\mathbf{k}_\tau^{(h)}
=
\mathbf{W}_{k}^{(h)}\mathbf{z}_\tau,\quad
\mathbf{v}_\tau^{(h)}
=
\mathbf{W}_{v}^{(h)}\mathbf{z}_\tau.
\end{equation}
Here, $\mathbf{q}_\tau^{(h)},\mathbf{k}_\tau^{(h)}\in\mathbb{R}^{d_k}$ and $\mathbf{v}_\tau^{(h)}\in\mathbb{R}^{d_v}$ are internal attention features. We omit the layer index for simplicity.

At step $t$, the Transformer attends over the context window $\mathcal{C}_t=\{t-L_t+1,\ldots,t\}$. The current query is compared with every key in this window:
\begin{equation}
\alpha_{t,\tau}^{(h)}
=
\operatorname{softmax}_{\tau\in\mathcal{C}_t}
\left(
\frac{
\mathbf{q}_{t}^{(h)\top}\mathbf{k}_{\tau}^{(h)}
}{
\sqrt{d_k}
}
\right),
\end{equation}
and the corresponding head output is
\begin{equation}
\mathbf{o}_{t}^{(h)}
=
\sum_{\tau\in\mathcal{C}_t}
\alpha_{t,\tau}^{(h)}
\mathbf{v}_{\tau}^{(h)}.
\end{equation}
The outputs of all heads are combined and passed through the subsequent backbone layers to obtain $\mathbf{y}_t$. In this formulation, historical information is accessed through explicit interactions between queries and the tokens stored in the context window. Consequently, computing self-attention over a sequence requires a quadratic number of token interactions with respect to the context length.

\subsubsection{Linear-attention backbone.}
LANav instead uses linear attention to integrate navigation history through recurrent state updates. Rather than retaining historical tokens and repeatedly computing attention over them, each attention head maintains a state matrix $\mathbf{S}_t^{(h)}\in\mathbb{R}^{d_k\times d_v}$ whose size is independent of the context length. At each navigation step, the state is updated using the current key-value pair:
\begin{equation}
\mathbf{S}_t^{(h)}
=
\operatorname{Update}
\left(
\mathbf{S}_{t-1}^{(h)},
\mathbf{k}_t^{(h)},
\mathbf{v}_t^{(h)}
\right).
\end{equation}
The current query then reads from the updated state:
\begin{equation}
\mathbf{o}_t^{(h)}
=
\mathbf{q}_t^{(h)\top}
\mathbf{S}_t^{(h)}.
\end{equation}
Thus, the Transformer retrieves historical information from an explicit token sequence, whereas linear attention compresses past key-value information into a recurrent state and updates this state as new observations arrive. This replaces full-window pairwise attention with incremental state updates, making the policy naturally compatible with sequential navigation.

Within this framework, different linear-attention backbones mainly differ in how $\mathbf{S}_t^{(h)}$ is updated. Linear Transformer~\cite{schlag2021linear} directly accumulates key-value associations into the state. DeltaNet~\cite{deltanet} updates the state according to the prediction error of the existing memory, with a learned write coefficient controlling the correction strength. Gated DeltaNet~\cite{yanggated} further introduces a data-dependent retention gate to regulate how much of the previous state is preserved. Detailed formulations of these existing update rules are provided in the \textit{Supplementary Material S1}.

These differences are particularly important for OVON. During navigation, consecutive egocentric observations can be highly redundant, goal-relevant objects or scene cues may be visible for only a few steps, and previously observed evidence may lose relevance as the agent moves through the environment. Therefore, effectively modeling navigation history requires more than accumulating a longer sequence of observations. The policy must determine how historical evidence is written into memory, how long it should be retained, and how it is used for current action prediction. To better model these processes, we further introduce WSLA, which maintains multiple linear-attention sub-states and combines their outputs through a learned weighted readout.

\begin{table*}[!t]
\centering
\caption{\textbf{Policy backbone comparison on HM3D-OVON.}
Reproduced models follow the same DAgRL~\cite{ovon} setting except for the policy backbone, and results are mean$\pm$std over three seeds. EALM~\cite{zemskova2025ovsegdtsegmentingtransformeropenvocabulary} uses a different training strategy and reports official results. Attention-based reproduced models use $l=500$.
Avg is the average SR over three splits; EALM is marked N/A because VAL SEEN SYN is not reported.}
\label{tab:policy_backbone}

\setlength{\tabcolsep}{3.2pt}
\renewcommand{\arraystretch}{1.05}

\resizebox{\textwidth}{!}{%
\begin{tabular}{llccccccc}
\toprule
Category & Method
 & \multicolumn{2}{c}{VAL SEEN}
 & \multicolumn{2}{c}{VAL SEEN SYN}
 & \multicolumn{2}{c}{VAL UNSEEN}
 & Avg \\
 &
 & SR($\uparrow$) & SPL($\uparrow$)
 & SR($\uparrow$) & SPL($\uparrow$)
 & SR($\uparrow$) & SPL($\uparrow$)
 & SR($\uparrow$) \\
\midrule

\multirow{1}{*}{RNN-based}
& GRU
 & 39.0$_{\pm0.3}$ & 15.0$_{\pm0.3}$
 & 29.3$_{\pm0.3}$ & 11.5$_{\pm0.1}$
 & 11.7$_{\pm0.3}$ & 3.8$_{\pm0.1}$
 & 26.7$_{\pm0.2}$\\

\midrule

\multirow{3}{*}{Transformer-based}
& Transformer~\cite{touvron2023llamaopenefficientfoundation}
 & 41.5$_{\pm0.3}$ & 21.0$_{\pm0.3}$
 & 29.9$_{\pm0.3}$ & 14.5$_{\pm0.1}$
 & 19.0$_{\pm0.3}$ & 7.1$_{\pm0.1}$
 & 30.1$_{\pm0.2}$ \\

& DAgRL~\cite{ovon}
 & 41.3$_{\pm0.3}$ & 21.2$_{\pm0.3}$
 & 29.4$_{\pm0.3}$ & 14.4$_{\pm0.1}$
 & 18.3$_{\pm0.3}$ & 7.9$_{\pm0.1}$
 & 29.7$_{\pm0.2}$ \\

& EALM~\cite{zemskova2025ovsegdtsegmentingtransformeropenvocabulary}
 & 42.5$_{\pm0.4}$ & 21.3$_{\pm0.2}$
 & N/A & N/A
 & 20.2$_{\pm0.5}$ & \textbf{8.8}$_{\pm0.2}$
 & N/A \\

\midrule

\multirow{4}{*}{LANav (ours)}
& Linear Transformer~\cite{schlag2021linear}
 & 42.9$_{\pm0.5}$ & 18.8$_{\pm0.4}$
 & 31.8$_{\pm0.4}$ & 12.9$_{\pm0.2}$
 & 19.2$_{\pm0.3}$ & 7.1$_{\pm0.1}$
 & 31.3$_{\pm0.2}$ \\

& DeltaNet~\cite{deltanet}
 & 45.3$_{\pm0.5}$ & 21.2$_{\pm0.4}$
 & 35.0$_{\pm0.4}$ & 15.1$_{\pm0.2}$
 & 21.5$_{\pm0.4}$ & 7.5$_{\pm0.3}$
 & 33.9$_{\pm0.3}$ \\

& Gated DeltaNet~\cite{yanggated}
 & 46.6$_{\pm0.3}$ & 22.4$_{\pm0.3}$
 & 36.6$_{\pm0.4}$ & 15.5$_{\pm0.2}$
 & 20.5$_{\pm0.3}$ & 7.5$_{\pm0.2}$
 & 34.6$_{\pm0.2}$ \\

& WSLA (ours)
 & \textbf{47.8}$_{\pm0.4}$ & \textbf{22.9}$_{\pm0.3}$
 & \textbf{39.8}$_{\pm0.5}$ & \textbf{16.2}$_{\pm0.3}$
 & \textbf{21.6}$_{\pm0.4}$ & 7.9$_{\pm0.3}$
 & \textbf{36.4}$_{\pm0.3}$ \\

\bottomrule
\end{tabular}%
}
\end{table*}



\subsubsection{WSLA}

To enhance the expressivity of LA, we propose WSLA.
Building upon the head-wise state expansion mechanism~\cite{liu2025scaling}
to increase state capacity, WSLA further introduces
a learnable weighting mechanism over expanded sub-heads,
enabling the model to regulate the readout contributions of expanded sub-states.
The \textit{Analysis of WSLA} subsection analyzes the effect of this design.

Specifically, we make the head index explicit and denote
the per-head recurrent state of the gated linear-attention backbone as
$\mathbf{S}_t^{(h)} \in \mathbb{R}^{d_k \times d_v}$.
WSLA increases intra-head capacity by expanding each head
into $E$ independent sub-head states.
For each of the $H$ original heads $h$,
linear projections expand the corresponding query and key vectors
into $E$ independent sub-heads:
\begin{align}
\tilde{\mathbf{q}}_t^{(h)}
&= \mathbf{W}_{q,h}^{\mathrm{exp}}\mathbf{q}_t^{(h)}
\in \mathbb{R}^{E \times d_k},
\\
\tilde{\mathbf{k}}_t^{(h)}
&= \mathbf{W}_{k,h}^{\mathrm{exp}}\mathbf{k}_t^{(h)}
\in \mathbb{R}^{E \times d_k}.
\end{align}
where $\mathbf{W}_{q,h}^{\mathrm{exp}}, \mathbf{W}_{k,h}^{\mathrm{exp}}
\in \mathbb{R}^{(E \cdot d_k)\times d_k}$ are expansion matrices.
We denote the $e$-th sub-head vectors as
$\mathbf{q}_t^{(h,e)}$ and $\mathbf{k}_t^{(h,e)}$.
To maintain parameter efficiency,
the value vector $\mathbf{v}_t^{(h)}$ is replicated
and shared across sub-heads.

Each sub-head $(h,e)$ maintains its own state
$\mathbf{S}_t^{(h,e)} \in \mathbb{R}^{d_k \times d_v}$.
In WSLA, the gating factors are predicted independently
for each sub-head, yielding per-sub-head gates
$\beta_t^{(h,e)},\gamma_t^{(h,e)}\in(0,1)$ using the same
gate parameterization as Gated DeltaNet.

For each sub-head, we first decay the previous state and
compute the prediction error with respect to the retained state:
\begin{equation}
\begin{aligned}
\widetilde{\mathbf{S}}_{t-1}^{(h,e)}
&=
\gamma_t^{(h,e)}
\mathbf{S}_{t-1}^{(h,e)},\\
\Delta\mathbf{v}_t^{(h,e)}
&=
\mathbf{v}_t^{(h)}
-
\left(
\widetilde{\mathbf{S}}_{t-1}^{(h,e)}
\right)^{\top}
\mathbf{k}_t^{(h,e)} .
\end{aligned}
\label{eq:wsla_delta}
\end{equation}
The gated delta update is then applied to each sub-head state:
\begin{equation}
\mathbf{S}_t^{(h,e)}
=
\widetilde{\mathbf{S}}_{t-1}^{(h,e)}
+
\beta_t^{(h,e)}\,
\mathbf{k}_t^{(h,e)}
\Delta\mathbf{v}_t^{(h,e)\top}.
\label{eq:wsla_update}
\end{equation}
The sub-head readout is then:
\begin{equation}
\mathbf{o}_t^{(h,e)}
=
\mathbf{q}_t^{(h,e)\top}
\mathbf{S}_{t}^{(h,e)}
\in \mathbb{R}^{d_v}.
\label{eq:wsla_readout}
\end{equation}

While state expansion enhances intra-head capacity, different sub-heads tend to capture heterogeneous aspects of historical observations. We therefore use a learnable logit vector $\boldsymbol{\lambda} \in \mathbb{R}^E$ shared across heads and time to regulate their readout contributions, with $\lambda^{(e)}$ denoting its $e$-th entry. The output of head $h$ is a softmax-normalized weighted combination of its expanded sub-heads:
\begin{equation}
\mathbf{u}_t^{(h)} = \sum_{e=1}^E \frac{\exp(\lambda^{(e)})}{\sum_{j=1}^E \exp(\lambda^{(j)})} \mathbf{o}_t^{(h,e)} \in \mathbb{R}^{d_v}
\end{equation}

The head outputs
$\{\mathbf{u}_t^{(h)}\}_{h=1}^{H}$
are concatenated as
$\mathbf{u}_t = \mathrm{Concat}(\mathbf{u}_t^{(1)}, \dots, \mathbf{u}_t^{(H)})$,
which is then RMS-normalized and modulated by an
input-dependent SiLU gate:
\begin{equation}
\overline{\mathbf{u}}_t
=
\operatorname{RMSNorm}(\mathbf{u}_t)
\odot
\operatorname{SiLU}(\mathbf{W}_g\mathbf{z}_t).
\end{equation}
The resulting gated representation replaces the attention
output in each backbone layer.

In our implementation, we use $H=4$ and $E=4$ in WSLA,
which maintains a parameter count and computational cost
(FLOPs) comparable to the other backbones.

\begin{figure*}[t]
\centering
\begin{minipage}[t]{0.32\textwidth}
    \centering
    \includegraphics[width=\textwidth]{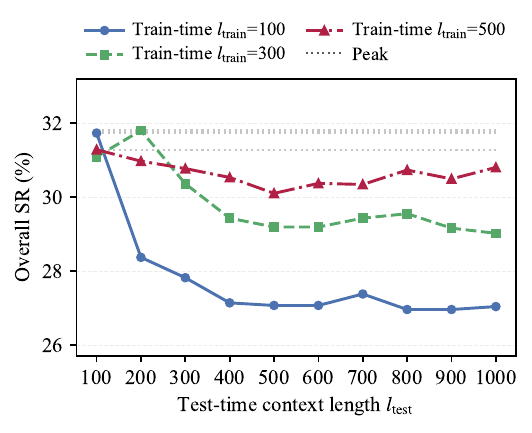}

    \caption*{(a) Transformer}
\end{minipage}
\hfill
\begin{minipage}[t]{0.32\textwidth}
    \centering
    \includegraphics[width=\textwidth]{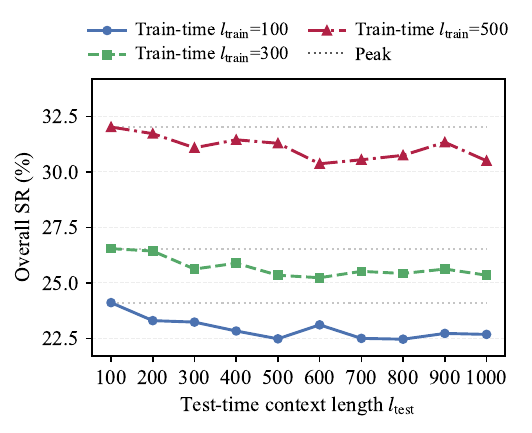}

    \caption*{(b) Linear Transformer}
\end{minipage}
\hfill
\begin{minipage}[t]{0.32\textwidth}
    \centering
    \includegraphics[width=\textwidth]{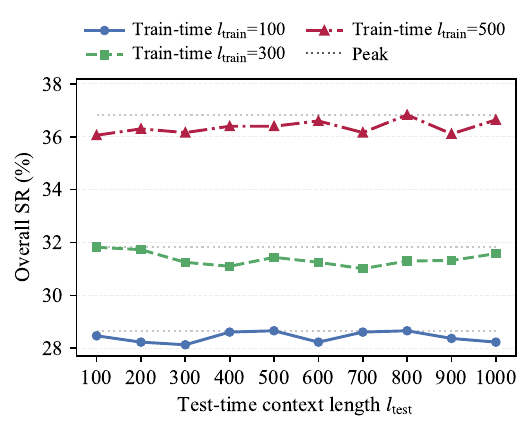}

    \caption*{(c) WSLA}
\end{minipage}

\caption{\textbf{Effect of context length.}
Overall SR (averaged across validation splits)
as a function of test-time context length $l_{\text{test}}$.
Each curve corresponds to a model trained with
$l_{\text{train}} \in \{100, 300, 500\}$.}
\label{fig:context_scaling}

\end{figure*}




\section{Simulation Experiment}

\subsection{Experimental Settings}

\paragraph{Benchmarks}
We evaluate different methods on HM3D-OVON~\cite{ovon}, which provides three validation splits for semantic generalization:

\begin{itemize}
    \item {VAL SEEN}: goal categories observed during training,
    \item {VAL SEEN SYN}: unseen but semantically similar goal names,
    \item {VAL UNSEEN}: semantically distinct unseen categories.
\end{itemize}

\paragraph{Evaluation Metrics}
We report Success Rate (SR) and Success weighted by Path Length (SPL)~\cite{objectnav}. SR measures task completion, while SPL additionally penalizes inefficient exploration.

\paragraph{Training Details}
We follow the two-stage training pipeline of DAgRL~\cite{ovon}. In the first stage, policies are pre-trained for 150M steps using DAgger~\cite{ross2011reductionimitationlearningstructured}. In the second stage, policies are fine-tuned for an additional 100M steps using PPO, initialized from the DAgger pre-trained checkpoint.

All methods are trained across 8 NVIDIA A100 80GB GPUs with 32 parallel environments per GPU, utilizing Variable Experience Rollout~\cite{wijmans2022verscalingonpolicyrl}, where the rollout length is set to match the context length $l$, alongside 16 mini-batches per update. We use Adam with learning rate \(2.5\times 10^{-4}\), \(\epsilon=10^{-5}\) and gradient clipping at \(0.2\). The code will be made publicly available to facilitate reproducibility.

\subsection{Linear Attention vs. Transformer}

We compare RNN-, Transformer-, and LA-based policy backbones on
HM3D-OVON under matched settings whenever the models are reproduced.
This comparison isolates the temporal policy backbone while keeping the
visual encoder, language encoder, training pipeline, and optimization
settings fixed.

As shown in Table~\ref{tab:policy_backbone}, LA-based backbones
consistently achieve higher SR than RNN- and Transformer-based baselines.
The advantage is clear on VAL SEEN SYN, indicating stronger
generalization to semantically related but unseen goal names. Within the
LA family, performance improves from Linear Transformer to DeltaNet,
Gated DeltaNet, and finally WSLA, suggesting that more structured and
regulated state updates lead to better navigation performance.

\subsection{Analysis of Linear Attention in OVON}\label{ana_la}

To understand why LA improves navigation performance in OVON, we analyze its behavior from three complementary perspectives: context-length scaling, state update mechanisms, and distance-stratified navigation performance.

\subsubsection{Effect of Context Length}
We examine whether temporal backbones can exploit longer observation histories
in OVON. Policies use training context lengths of 100, 300, or 500 and test
context lengths from 100 to 1000 in increments of 100. Only the length used to
construct $\mathbf{z}_{t-(l-1):t}$ varies, and we report overall SR averaged
across all validation splits.

As shown in Fig.~\ref{fig:context_scaling}, Transformer policies do not benefit
from larger training contexts and can even degrade when
$l_{\text{train}}=500$, suggesting limited ability to exploit extended histories
under this setting. In contrast, LA backbones achieve higher peak SR, improve
consistently with larger $l_{\text{train}}$, and remain stable across different
$l_{\text{test}}$ values. These results indicate that LA provides a more
effective and robust mechanism for integrating historical observations in OVON.

\subsubsection{State Update Mechanism}

We further examine how state update design influences navigation performance.
As shown in Table~\ref{tab:policy_backbone}, LA-based backbones consistently
outperform both RNN-based and Transformer-based backbones under matched
settings. However, the vanilla Linear Transformer brings only marginal
improvements over the standard Transformer, suggesting that replacing
quadratic self-attention with a linear formulation alone is not sufficient.
The gains become more pronounced as the update rule becomes more structured
and regulated, as evidenced by the progressive improvements from Linear
Transformer to DeltaNet, Gated DeltaNet, and finally WSLA. These results highlight the importance of state-update design, beyond simply increasing the available context length. 





\begin{table}[t]
\centering
\small
\caption{\textbf{Distance-stratified evaluation on HM3D-OVON.}
All evaluation episodes are grouped by the initial geodesic
distance from the start location to the nearest goal instance.
$\Delta$ denotes the absolute difference between WSLA and the
Transformer (WSLA $-$ Transformer).}
\begin{tabular*}{\columnwidth}{@{\extracolsep{\fill}}llccc}
\toprule
Distance & Metric & WSLA & Transformer & $\Delta$ \\
\midrule
0--8m  & SR  & 41.04 & 34.33 & +6.71 \\
       & SPL & 17.00 & 15.64 & +1.36 \\
\midrule
8--15m & SR  & 24.52 & 17.60 & +6.92 \\
       & SPL & 13.13 &  9.75 & +3.38 \\
\midrule
15m+   & SR  & 14.36 &  6.68 & +7.68 \\
       & SPL &  7.88 &  3.67 & +4.21 \\
\bottomrule
\end{tabular*}
\label{tab:distance}
\end{table}

\begin{table}[t]
\centering

\caption{\textbf{Ablation study of WSLA.}}
\begin{tabular*}{\columnwidth}{@{\extracolsep{\fill}}ccc}
\toprule
State Expansion & Weighted & SR $\uparrow$ \\
\midrule
 & & 34.6 \\
\checkmark & & 35.2 \\
\checkmark & \checkmark & \textbf{36.4} \\
\bottomrule
\end{tabular*}
\label{tab:ablation_wsla}
\end{table}

\begin{figure}[t]
    \centering
    \includegraphics[width=1\linewidth]{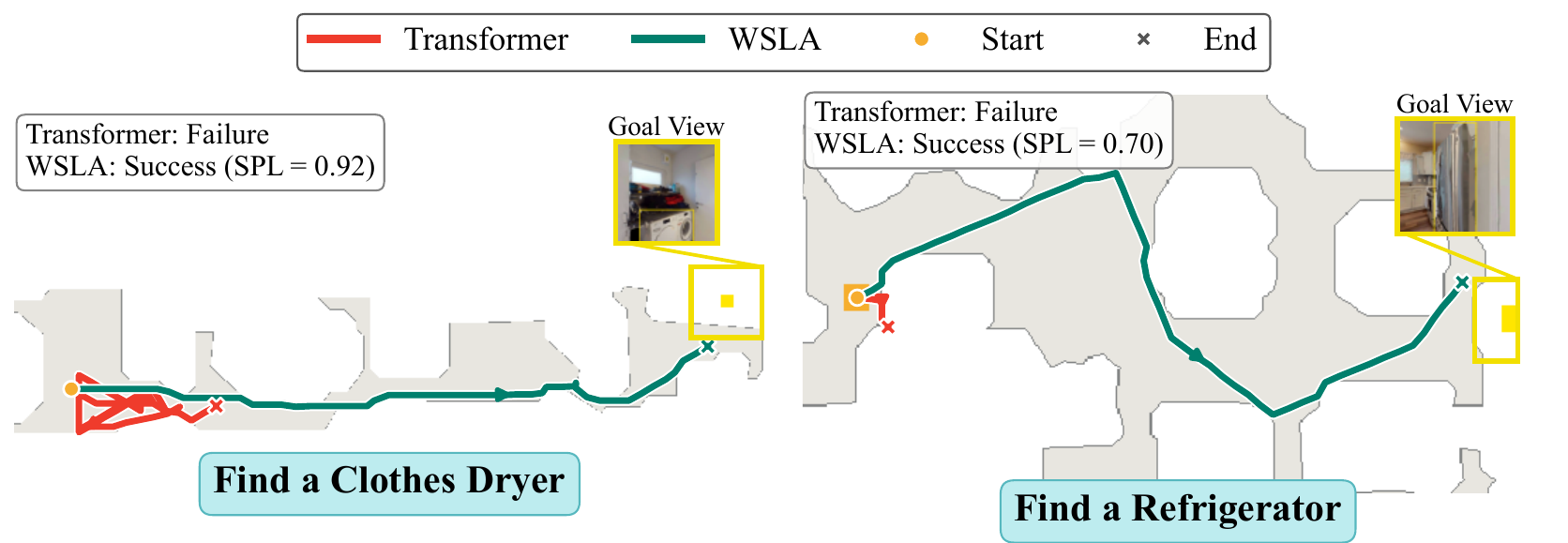}
    \caption{
    \textbf{Qualitative comparison of navigation behaviors on HM3D-OVON.}
    }
    \label{fig:trajectory}
\end{figure}

\begin{figure}[t]
    \centering
    \includegraphics[width=1\linewidth]{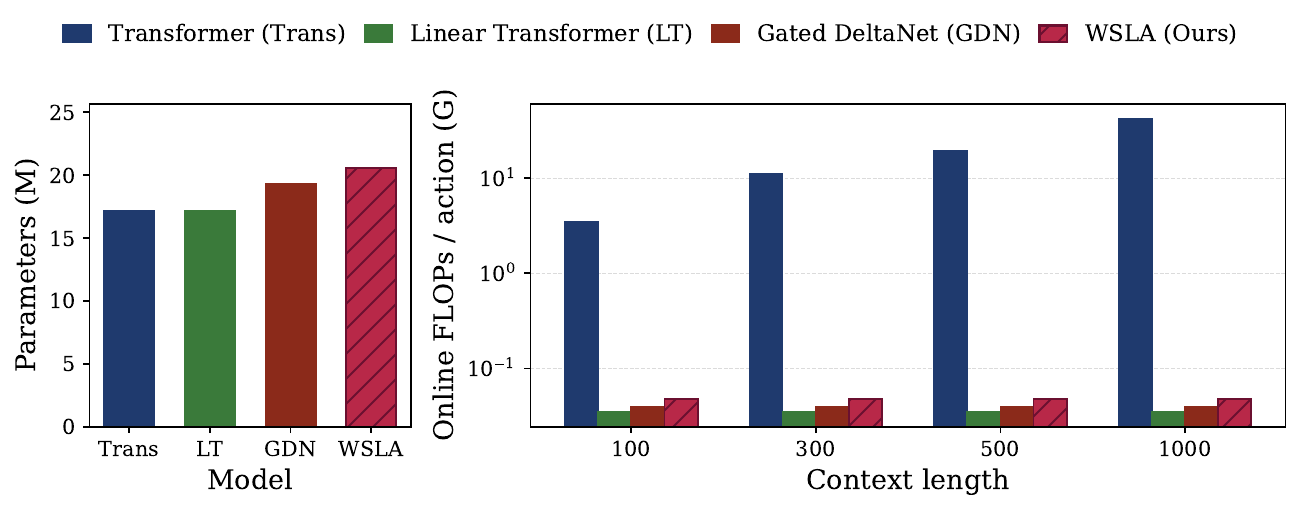}
    \caption{
    \textbf{Parameter count and computational cost.}
    Model parameters (left) and online FLOPs (right) of different policy backbones under identical architectural settings.
    }
    \label{fig:efficiency}
\end{figure}


\subsubsection{Distance-Based Evaluation.}
We group HM3D-OVON evaluation episodes by their initial geodesic distance to
the nearest goal instance and compare WSLA with the Transformer baseline in
each range.
As shown in Table~\ref{tab:distance}, WSLA consistently outperforms the Transformer across all distance bins.
Moreover, both SR and SPL gains increase from short- to long-distance episodes, suggesting that WSLA supports more effective and efficient navigation when extended exploration is required.
We provide a trajectory analysis of WSLA compared with the Transformer baseline in Fig.~\ref{fig:trajectory}.
These examples illustrate that, during inference, the Transformer either repeatedly explores a limited local region or terminates prematurely near the start, whereas WSLA sustains exploration over longer horizons and successfully reaches the target. Additional paired trajectory diagnostics are provided in the \textit{Supplementary Material S3}.

\subsubsection{Analysis of WSLA}
\label{ana_sela}

We conduct ablation studies to analyze the effects of state expansion and weighted aggregation in WSLA.
As shown in Table~\ref{tab:ablation_wsla}, expanding the state representation improves SR while keeping the parameter size nearly unchanged, indicating that increasing intra-head capacity enhances the expressivity of the policy backbone.
Furthermore, introducing learnable weights to regulate the contributions of expanded sub-heads leads to additional performance gains.
This suggests that sub-states contribute unequally to navigation decisions, and that learned weighting to emphasize informative sub-states improves navigation performance.



\subsection{Efficiency Analysis}
We compare model size and online FLOPs per action under identical
architectural settings. As shown in Fig.~\ref{fig:efficiency},
WSLA incurs only a moderate parameter increase from state expansion.
Transformer FLOPs rise from 3.52G at $l=100$ to 42.53G at
$l=1000$ because the full context is recomputed at every step.
LA backbones instead update cached states using only the current
token, keeping the cost nearly constant. WSLA remains around
0.048G FLOPs/action. WSLA incurs slightly higher online FLOPs per action than other LA variants, but remains considerably below the Transformer. Meanwhile, it achieves the best navigation performance, suggesting a favorable balance between effectiveness and online computational cost.





\subsection{Generalization and Adaptation to HSSD}

To examine whether WSLA generalizes beyond HM3D-OVON, we compare it with the Transformer on the Habitat Synthetic Scenes Dataset (HSSD) ObjectNav benchmark~\cite{khanna2024hssd}, whose different scene distribution provides a complementary test of dataset-level robustness. We evaluate zero-shot transfer and 60M-step HSSD fine-tuning with matched policy inputs, action space, context length, and metrics. As shown in Table~\ref{tab:hssd_generalization}, WSLA improves over the Transformer by 9.62 SR and 1.31 SPL in zero-shot transfer and by 8.33 SR and 10.17 SPL after fine-tuning. The larger SPL gain indicates more efficient adaptation, demonstrating WSLA's robustness to dataset shift and effectiveness beyond HM3D-OVON.

 \begin{table}[t]
\centering
\caption{\textbf{Generalization and adaptation results on HSSD.}}
\label{tab:hssd_generalization}

\begin{tabular*}{\columnwidth}{@{\extracolsep{\fill}}llcc}
\toprule
Setting & Method & SR $\uparrow$ & SPL $\uparrow$ \\
\midrule
\multirow{2}{*}{Zero-shot}
& Transformer & 19.55 & 7.13 \\
& WSLA (ours) & \textbf{29.17} & \textbf{8.44} \\
\midrule
\multirow{2}{*}{HSSD FT}
& Transformer & 56.17 & 23.50 \\
& WSLA (ours) & \textbf{64.50} & \textbf{33.67} \\
\bottomrule
\end{tabular*}
\end{table}

\subsection{Real-World Validation}

\begin{figure}[t]
    \centering
    \includegraphics[width=1\linewidth]{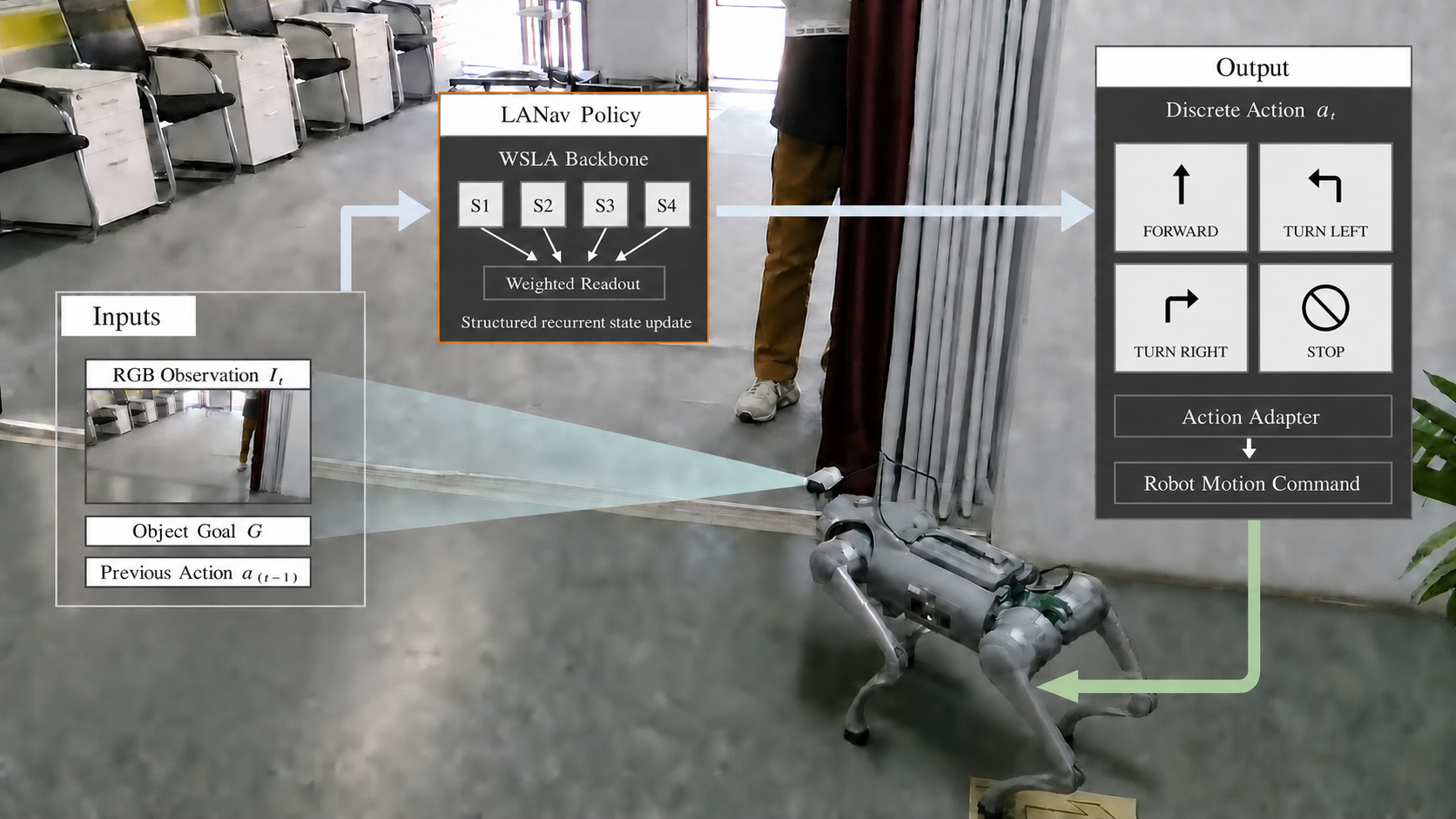}
    \caption{
    \textbf{Real-world validation.}
    Deployment of LANav with the WSLA backbone on a physical robot for open-vocabulary object-goal navigation.
    }
    \label{fig:real_world}
\end{figure}



To evaluate sim-to-real transfer, we deploy LANav with the WSLA backbone on a Unitree Go2 robot in an indoor environment, as shown in Fig.~\ref{fig:real_world}. The policy receives similar inputs as in HM3D-OVON: an egocentric RGB observation, a language-specified object goal, and the previous action. At each step, it predicts a discrete navigation action, which is converted into robot motion commands by a lightweight action adapter. 

We evaluate five object-goal categories, including a trash bin, box, plant, chair, and table, from two distinct starting locations. Each target-start pair is repeated five times, resulting in 50 trials. LANav succeeds in 41 trials, achieving an overall SR of 82\%. These results demonstrate the feasibility of deploying LANav on a physical robot and provide evidence of effective sim-to-real transfer. More implementation details and videos are provided in the \textit{Supplementary Material S3}.

\section{Conclusion}

We investigate LA as a policy backbone for OVON and propose LANav and WSLA, which use structured state updates and weighted sub-state aggregation. Under matched settings, LA backbones outperform RNN and Transformer baselines, benefit from longer training contexts, and remain stable across test-time lengths. These results highlight the importance of state-update design. WSLA achieves strong HM3D-OVON performance with favorable computational scaling, while HSSD and real-world evaluations demonstrate transferability. There are still limitations, including the use of shallow backbones and bounded context windows. Future work will scale model capacity and horizons and conduct broader real-world evaluation.

\bibliography{ref}

\clearpage
\setcounter{page}{1}
\setcounter{section}{0}
\setcounter{subsection}{0}
\setcounter{subsubsection}{0}
\setcounter{equation}{0}
\setcounter{table}{0}
\setcounter{figure}{0}
\setcounter{secnumdepth}{2}
\renewcommand{\thesection}{S\arabic{section}}
\renewcommand{\thesubsection}{\thesection.\arabic{subsection}}
\renewcommand{\thesubsubsection}{\thesubsection.\arabic{subsubsection}}
\numberwithin{equation}{section}
\renewcommand{\thetable}{S\arabic{table}}
\renewcommand{\thefigure}{S\arabic{figure}}

\twocolumn[
\begin{center}
{\LARGE\bfseries Supplementary Material\par}
\vspace{1.5em}
\end{center}
]

\section{Detailed Update Rules for Linear-Attention Baselines}
\label{sec:implementation}

This section provides the update rules of the baseline recurrent-state models used in our experiments. For all recurrence definitions, the associative state is initialized as $\mathbf{S}_0=\mathbf{0}$.

\paragraph{Linear Transformer.}
The kernelized baseline uses the positive feature map
$\phi(\mathbf{x})=\operatorname{ELU}(\mathbf{x})+1$ and maintains a key--value
accumulator $\mathbf{A}_t$ together with a normalizer $\mathbf{r}_t$:
\begin{align}
\mathbf{A}_t
&=\mathbf{A}_{t-1}+
\phi(\mathbf{k}_t)\mathbf{v}_t^{\top},
&
\mathbf{r}_t
&=\mathbf{r}_{t-1}+\phi(\mathbf{k}_t),
\label{eq:linear-states}
\\
\mathbf{o}_t
&=
\frac{\mathbf{A}_t^{\top}\phi(\mathbf{q}_t)}
{\max\!\left(\phi(\mathbf{q}_t)^{\top}\mathbf{r}_t,10^{-6}\right)}.
\label{eq:linear-readout}
\end{align}
Unlike the gated rules below, this recurrence contains no learned retention or
write coefficient.

\paragraph{DeltaNet.}
For each attention head $h$, the query and key projections in
Eq.~(7) are passed through a SiLU activation and independently
$\ell_2$-normalized. For notational simplicity, the resulting
vectors are still denoted by $\mathbf{q}_t^{(h)}$ and
$\mathbf{k}_t^{(h)}$. DeltaNet predicts a scalar write coefficient
from the current-layer token representation $\mathbf{z}_t$ and
computes the prediction error at the current key:
\begin{equation}
\begin{aligned}
\beta_t^{(h)}
&=
\sigma\!\left(
\mathbf{W}_{\beta}^{(h)}\mathbf{z}_t
\right),
\\
\widehat{\mathbf{v}}_t^{(h)}
&=
\left(\mathbf{S}_{t-1}^{(h)}\right)^{\top}
\mathbf{k}_t^{(h)},
\qquad
\Delta\mathbf{v}_t^{(h)}
=
\mathbf{v}_t^{(h)}
-
\widehat{\mathbf{v}}_t^{(h)} .
\end{aligned}
\end{equation}
Here,
$\mathbf{W}_{\beta}^{(h)}\in\mathbb{R}^{1\times d}$ and
$\beta_t^{(h)}\in(0,1)$ is the write coefficient of head $h$.
The state is then updated using the prediction-error correction:
\begin{equation}
\begin{aligned}
\mathbf{S}_t^{(h)}
&=
\mathbf{S}_{t-1}^{(h)}
+
\beta_t^{(h)}
\mathbf{k}_t^{(h)}
\left(\Delta\mathbf{v}_t^{(h)}\right)^{\top},
\\
\mathbf{o}_t^{(h)}
&=
\left(\mathbf{q}_t^{(h)}\right)^{\top}
\mathbf{S}_t^{(h)} .
\end{aligned}
\end{equation}
The coefficient $\beta_t^{(h)}$ controls how strongly the value
associated with the current key is corrected toward
$\mathbf{v}_t^{(h)}$. Unlike Gated DeltaNet, DeltaNet does not use
a separate state-retention coefficient.

\paragraph{Gated DeltaNet.}
For every token and head, a retention coefficient and a write coefficient are
predicted from the pre-normalized layer input:
\begin{equation}
\begin{aligned}
g_t
&=-\exp(A^{\log})\operatorname{softplus}
\!\left([\mathbf{W}_a\mathbf{z}_t]+b_{\Delta}\right),
\\
\gamma_t&=\exp(g_t),
\\
\beta_t
&=\sigma\!\left([\mathbf{W}_{\beta}\mathbf{z}_t]+b_{\beta}\right).
\end{aligned}
\label{eq:gdn-gates}
\end{equation}

Decay is applied before the prediction error is formed:
\begin{equation}
\begin{aligned}
\widetilde{\mathbf{S}}_{t-1}
&=\gamma_t\mathbf{S}_{t-1},
&
\Delta\mathbf{v}_t
&=\mathbf{v}_t-
\widetilde{\mathbf{S}}_{t-1}^{\top}\mathbf{k}_t,
\\
\mathbf{S}_t
&=\widetilde{\mathbf{S}}_{t-1}+
\beta_t\mathbf{k}_t\Delta\mathbf{v}_t^{\top}.
\end{aligned}
\label{eq:gdn-recurrence}
\end{equation}

After the recurrent readout, Gated DeltaNet applies the normalized
value-channel gate
\begin{equation}
\overline{\mathbf{o}}_t
=\operatorname{RMSNorm}(\mathbf{o}_t)
\odot\operatorname{SiLU}(\mathbf{W}_g\mathbf{z}_t).
\label{eq:gdn-output-gate}
\end{equation}
This gate affects the emitted representation but not the recurrent state in
Eq.~\eqref{eq:gdn-recurrence}.

\section{Additional Paired Trajectory Diagnostics}
\label{sec:trajectories}

The following figures provide additional qualitative trajectory comparisons
between Transformer and WSLA. Within each pair, the two policies receive the
same episode, initial pose, and goal, and act deterministically under a
500-step cap. Red and teal paths denote Transformer and WSLA, respectively;
orange circles mark the shared starting positions, crosses indicate the
terminal positions, and yellow annotations identify the goal-view locations.

\subsection{HM3D-OVON: Matched Success}
\label{sec:hm3d-matched-success}

\begin{figure}[t]
  \centering
  \includegraphics[width=\linewidth]{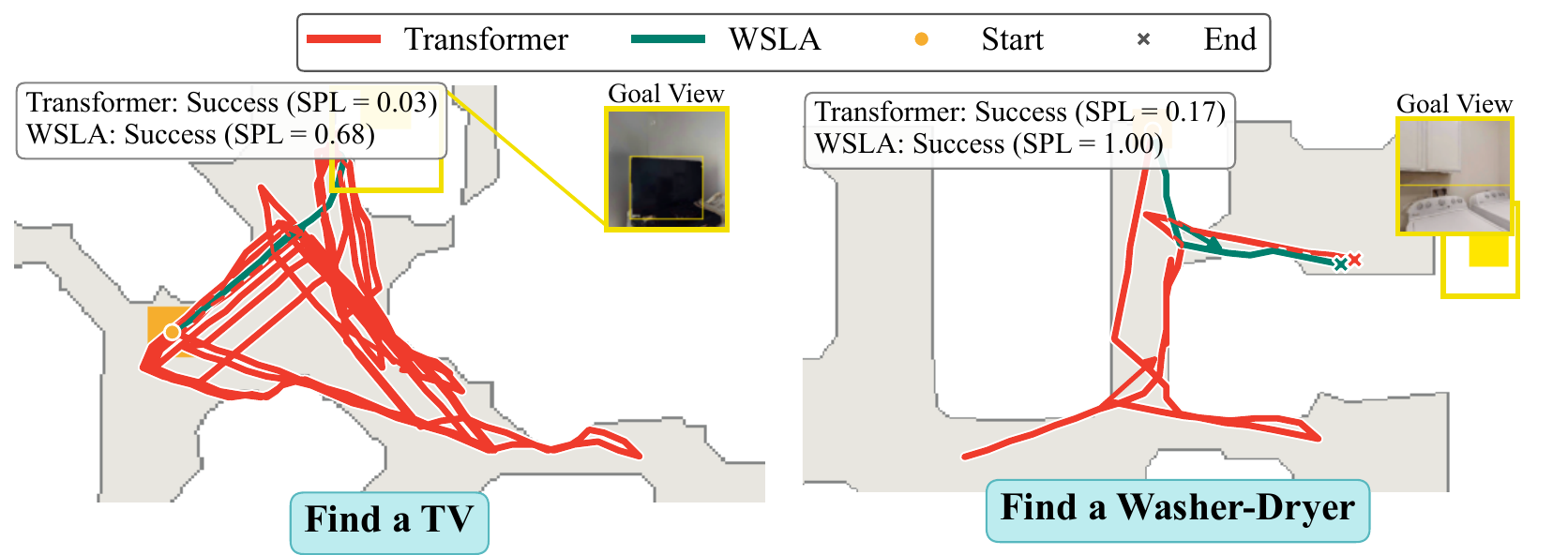}\par
  \includegraphics[width=\linewidth]{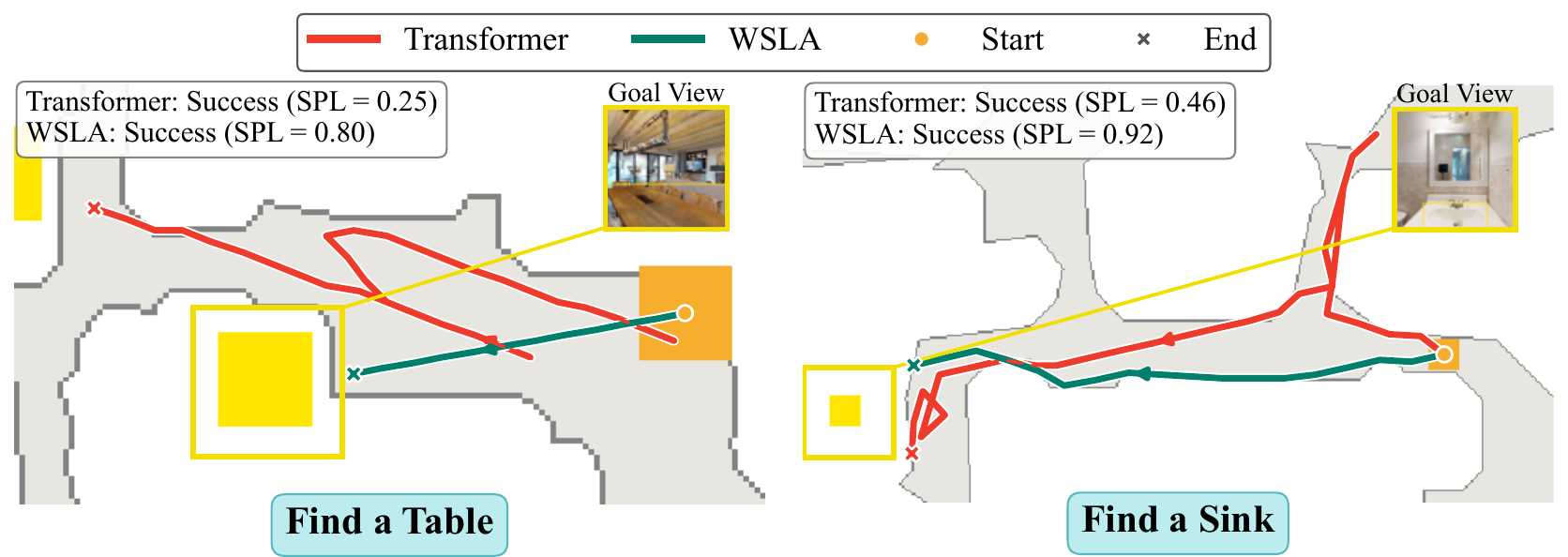}
  \caption{Additional HM3D-OVON VAL SEEN episodes.}
  \label{fig:hm3d-both-success}
\end{figure}

Figure~\ref{fig:hm3d-both-success} isolates route efficiency. In each
case, WSLA uses a shorter, less repetitive route. The most visually pronounced looping behavior occurs in the TV episode, where the Transformer repeatedly revisits the same local region.

\subsection{HSSD: Zero-Shot Transfer and Fine-Tuned Policies}
\label{sec:hssd-trajectories}

\begin{figure}[H]
  \centering
  \includegraphics[width=\linewidth]{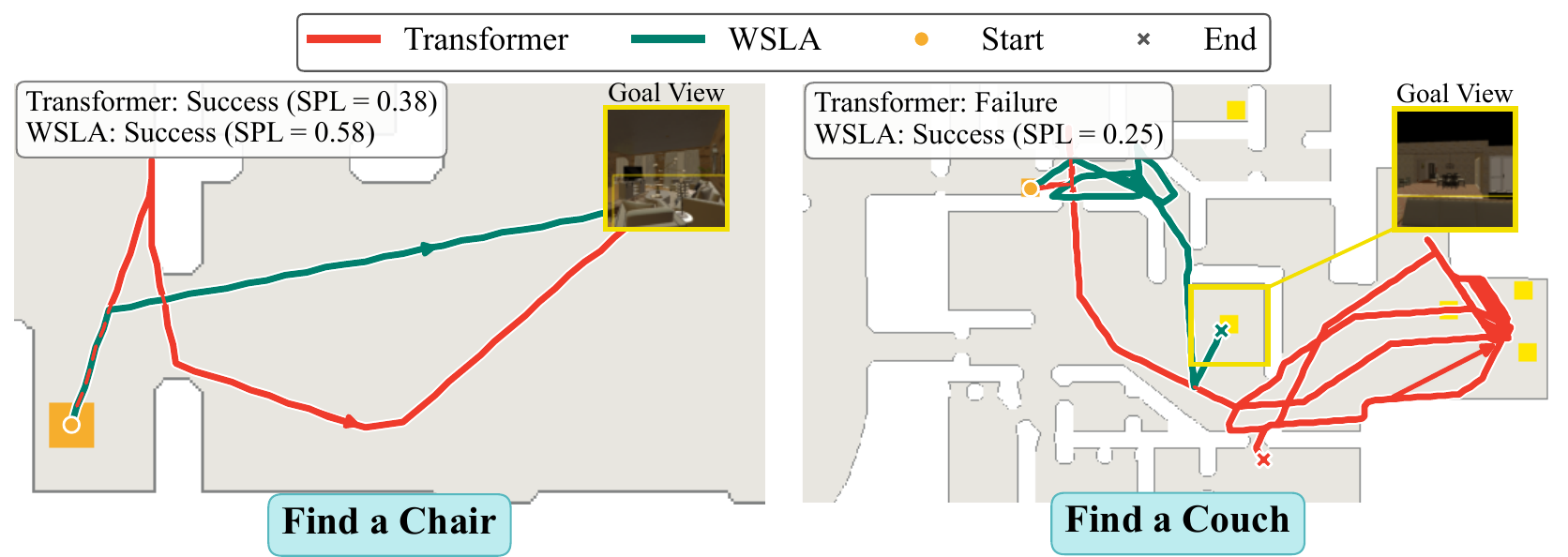}\par
  \vspace{-0.35em}
  \includegraphics[width=\linewidth]{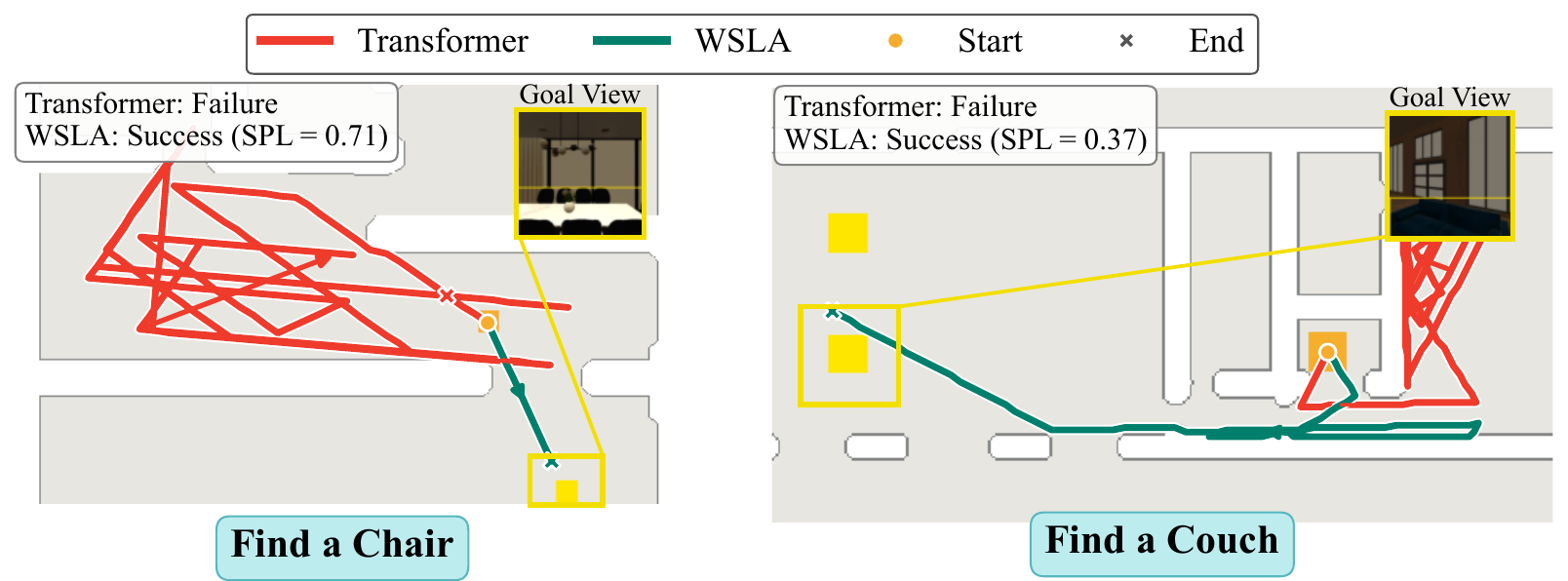}
  \caption{HSSD trajectory comparisons. Top: zero-shot transfer of HM3D-trained policies. Bottom: policies after matched HSSD fine-tuning. }
  \label{fig:hssd-trajectories}
\end{figure}

The zero-shot examples use HM3D-trained policies evaluated without HSSD
adaptation. The fine-tuned examples compare the two policies after matched
HSSD adaptation. Figure~\ref{fig:hssd-trajectories} places the two settings
in two compact rows.

\section{Real-World Deployment}
\label{sec:deployment}

This section records the hardware configuration, physical action interface,
state initialization, and visual evidence needed to interpret the real-world
experiment.

\subsection{Robot and Observation Interface}

The physical system uses a Unitree Go2 EDU equipped with an external Intel
RealSense D435i. The camera captures $640\times480$ RGB frames at 15 fps and is
mounted approximately 0.45 m above the ground with zero pitch. The policy
encodes the egocentric frames with SigLIP. The language goal remains fixed
within an episode, whereas the previous-action symbol changes after each
policy decision. At the beginning of every trial, the WSLA recurrent state is
reinitialized. External photographs and recordings serve as third-person
documentation of the setup and rollout.

\begin{figure}[t]
  \centering
  \includegraphics[height=0.42\columnwidth]{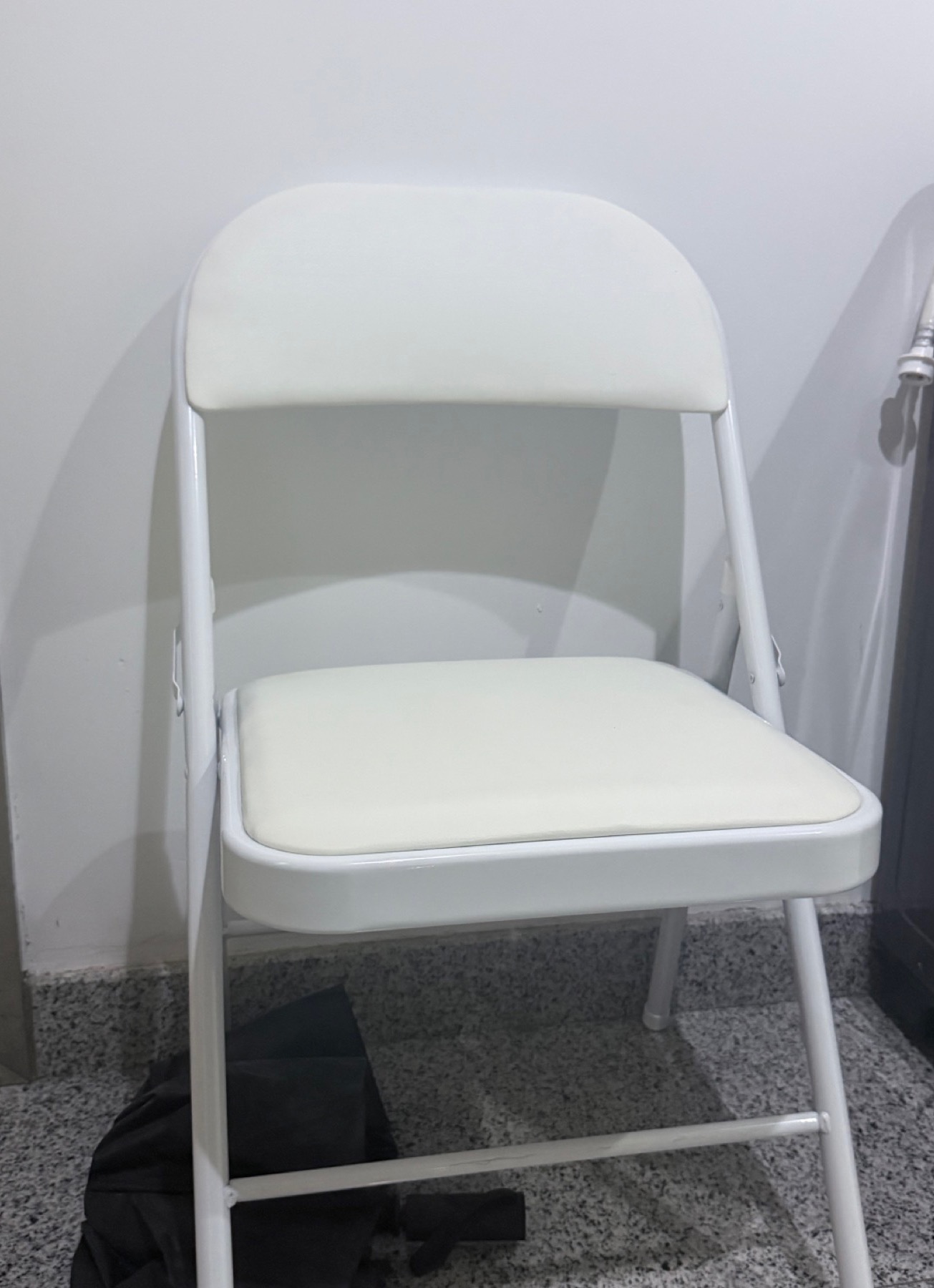}
  \includegraphics[height=0.42\columnwidth]{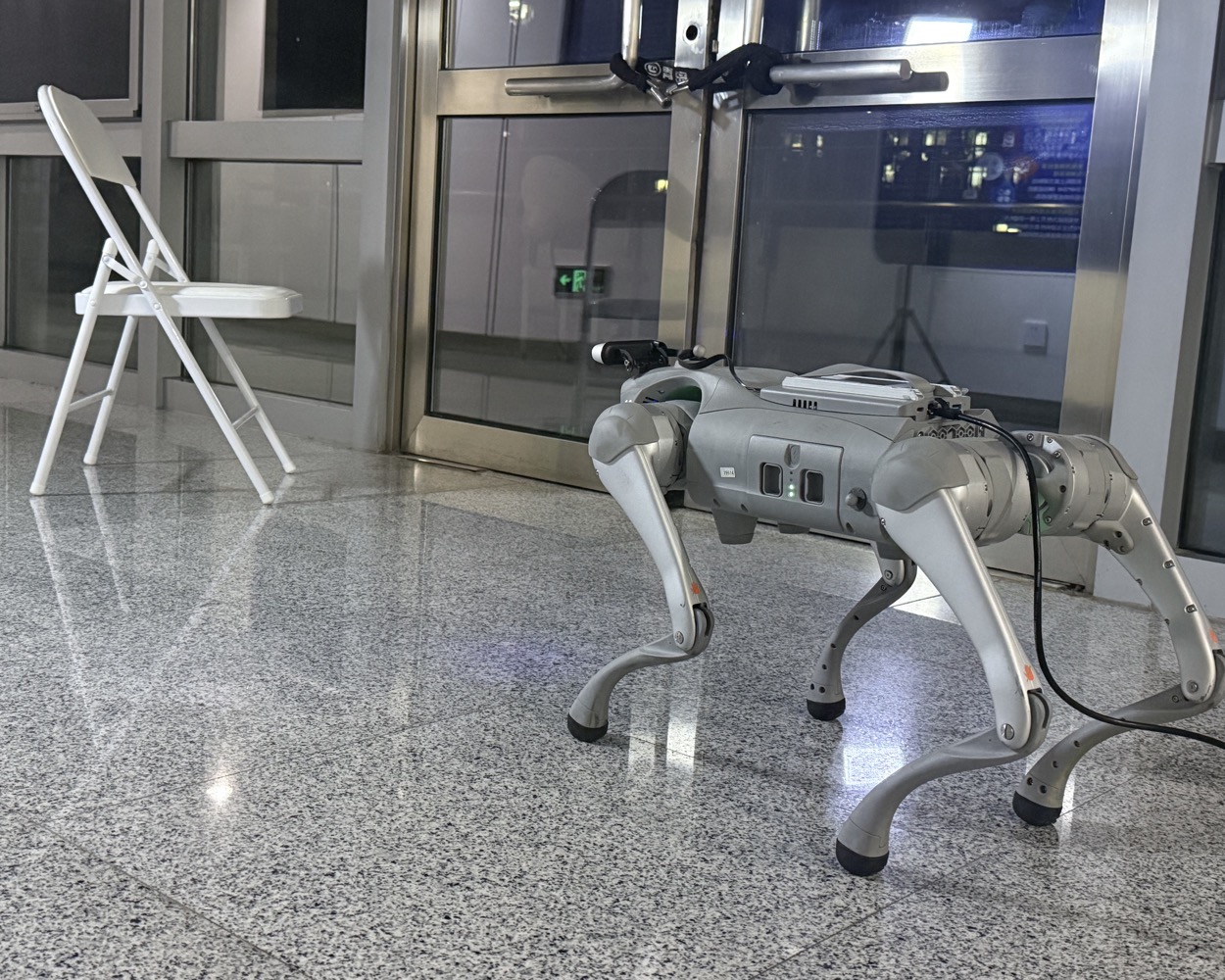}\par
  {\scriptsize (a) Chair}\par
  \includegraphics[height=0.42\columnwidth]{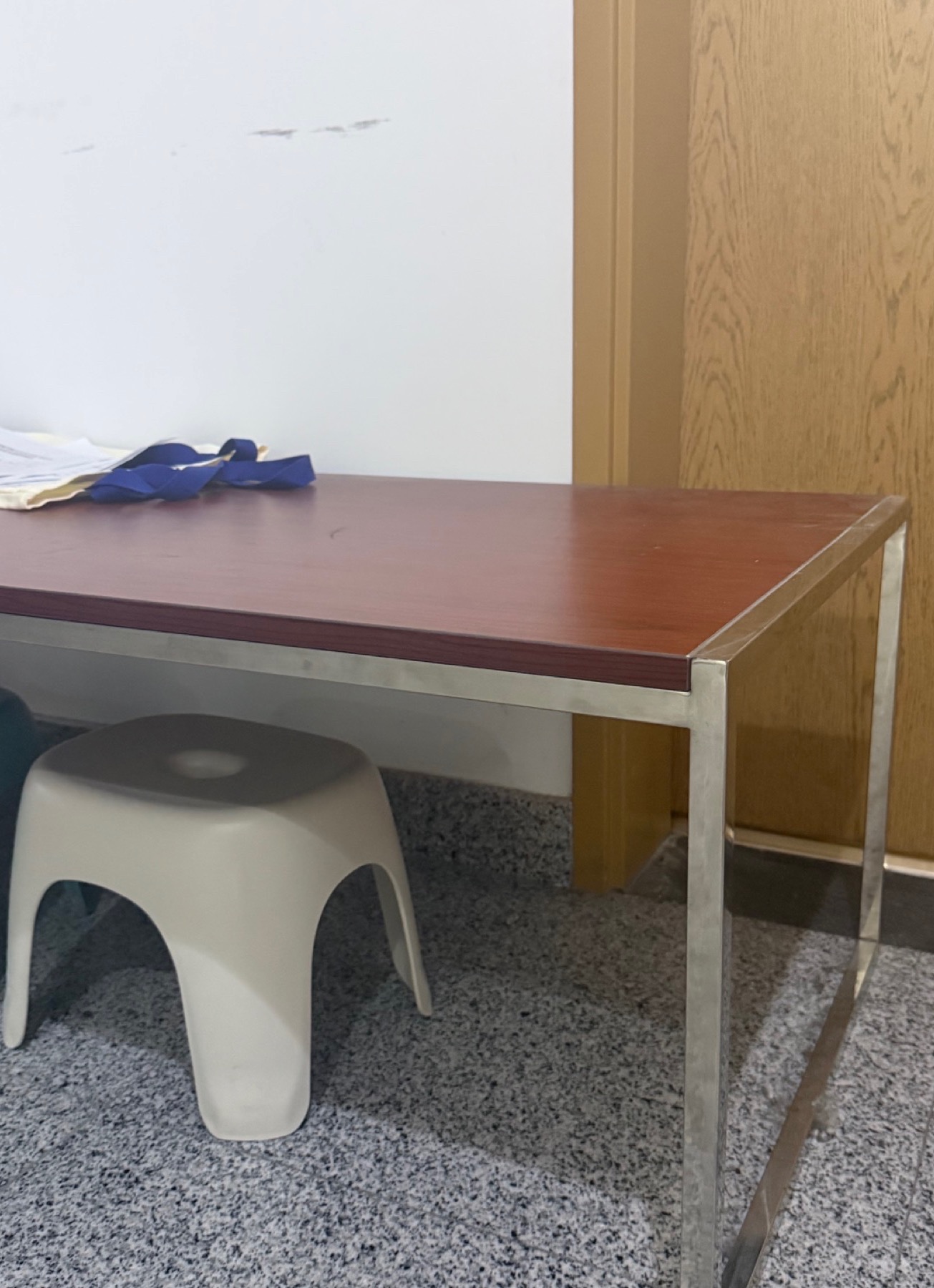}
  \includegraphics[height=0.42\columnwidth]{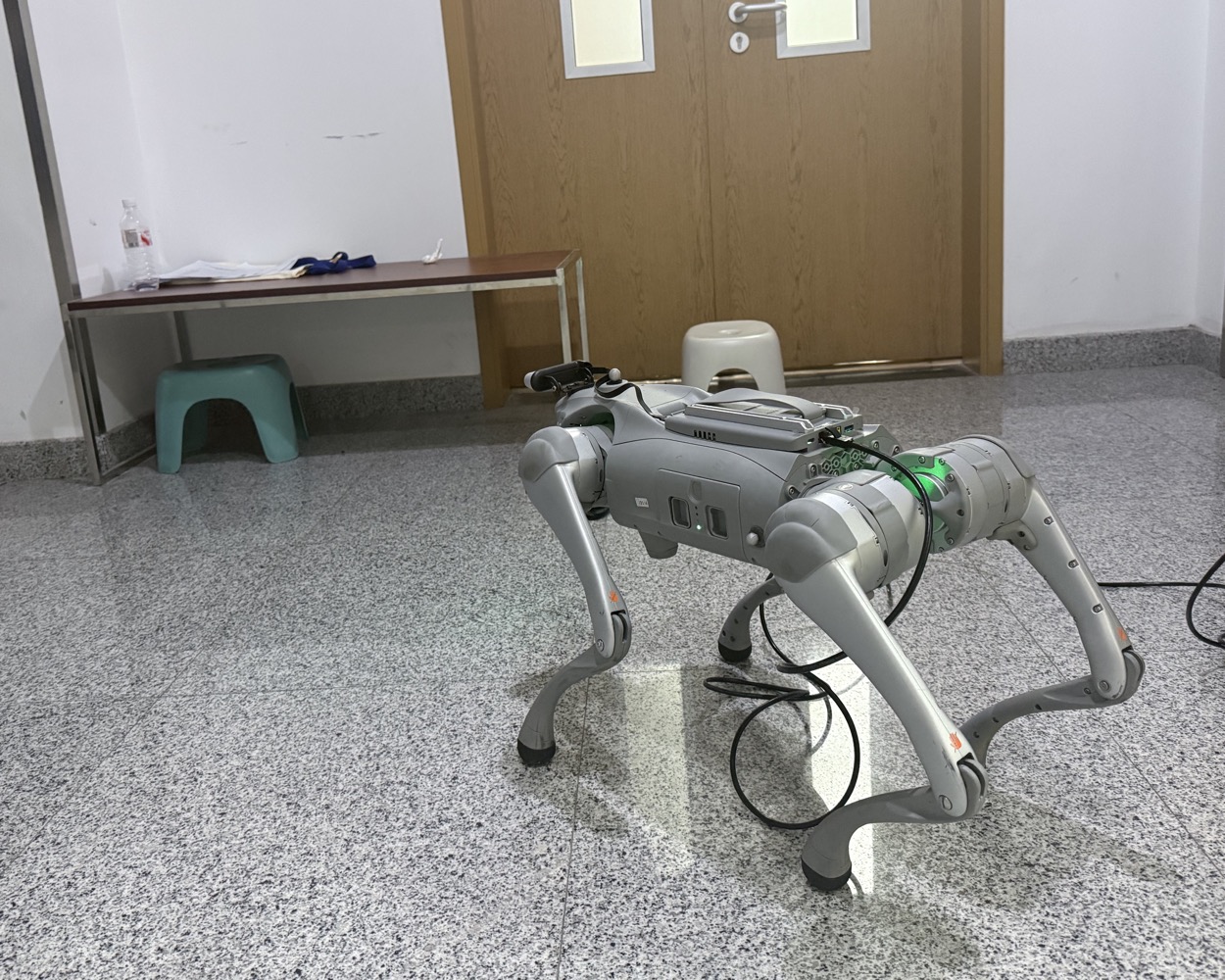}\par
  {\scriptsize (b) Table}\par
  \includegraphics[height=0.42\columnwidth]{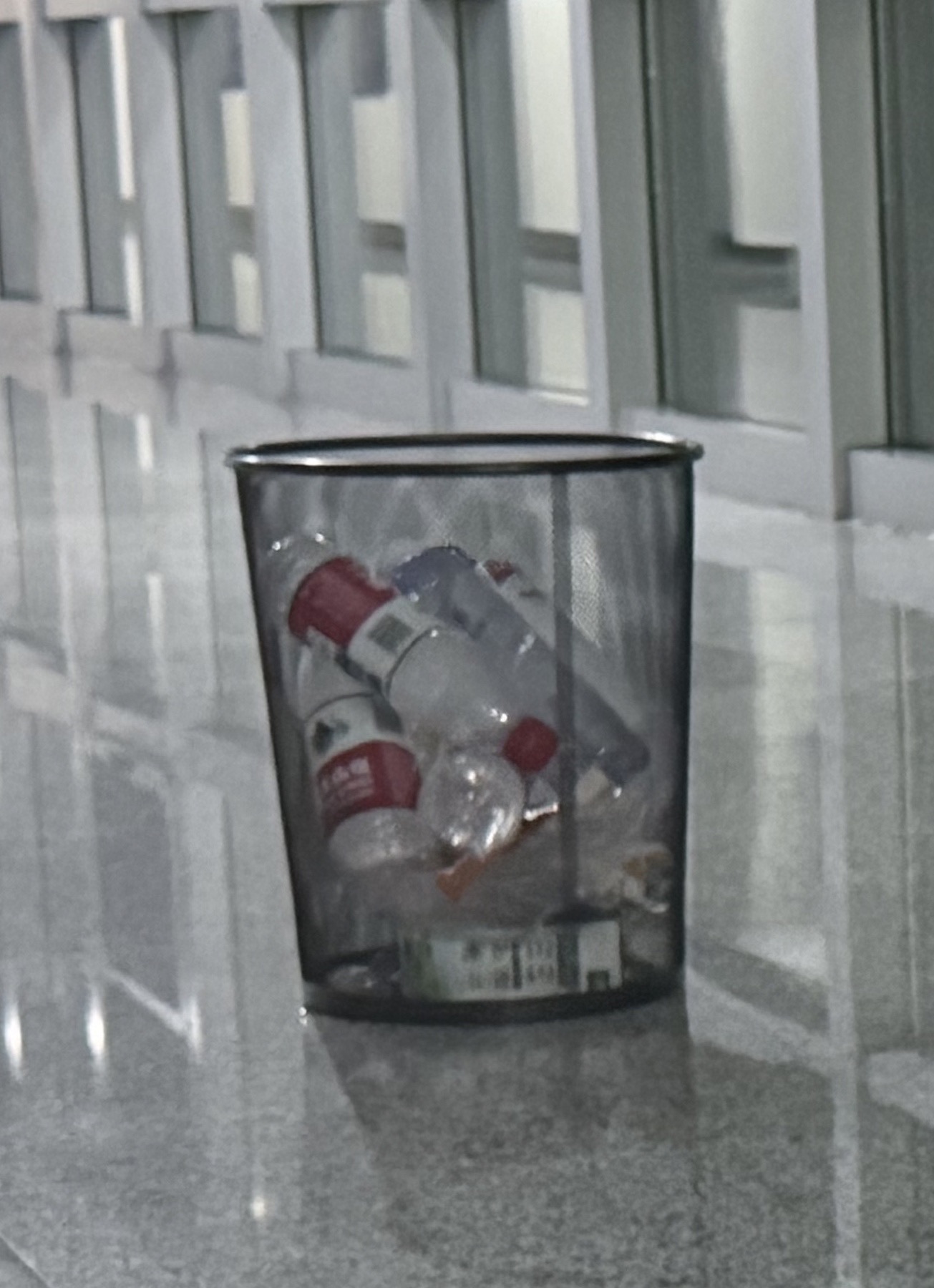}
  \includegraphics[height=0.42\columnwidth]{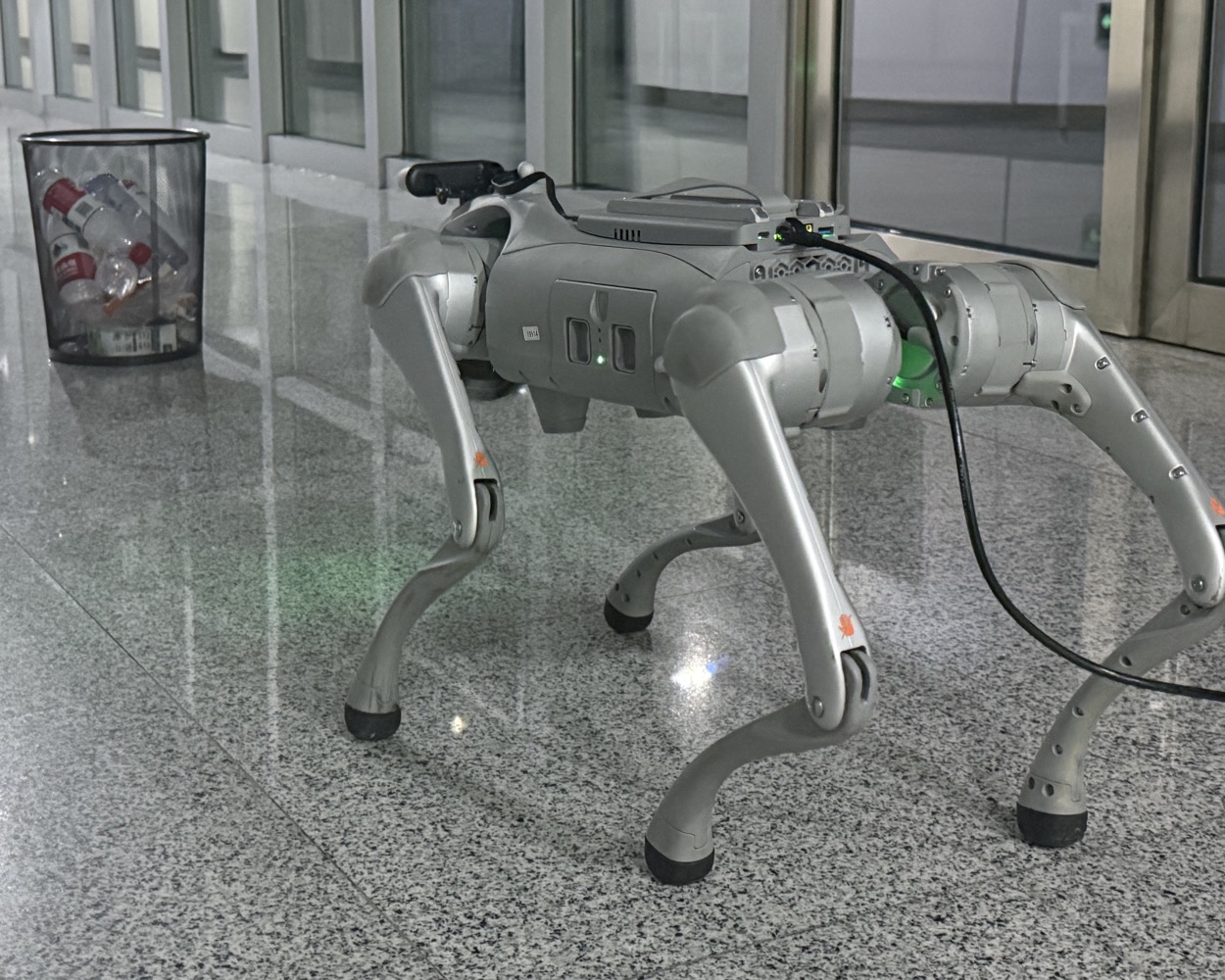}\par
  {\scriptsize (c) Trash Bin}\par
  \includegraphics[height=0.42\columnwidth]{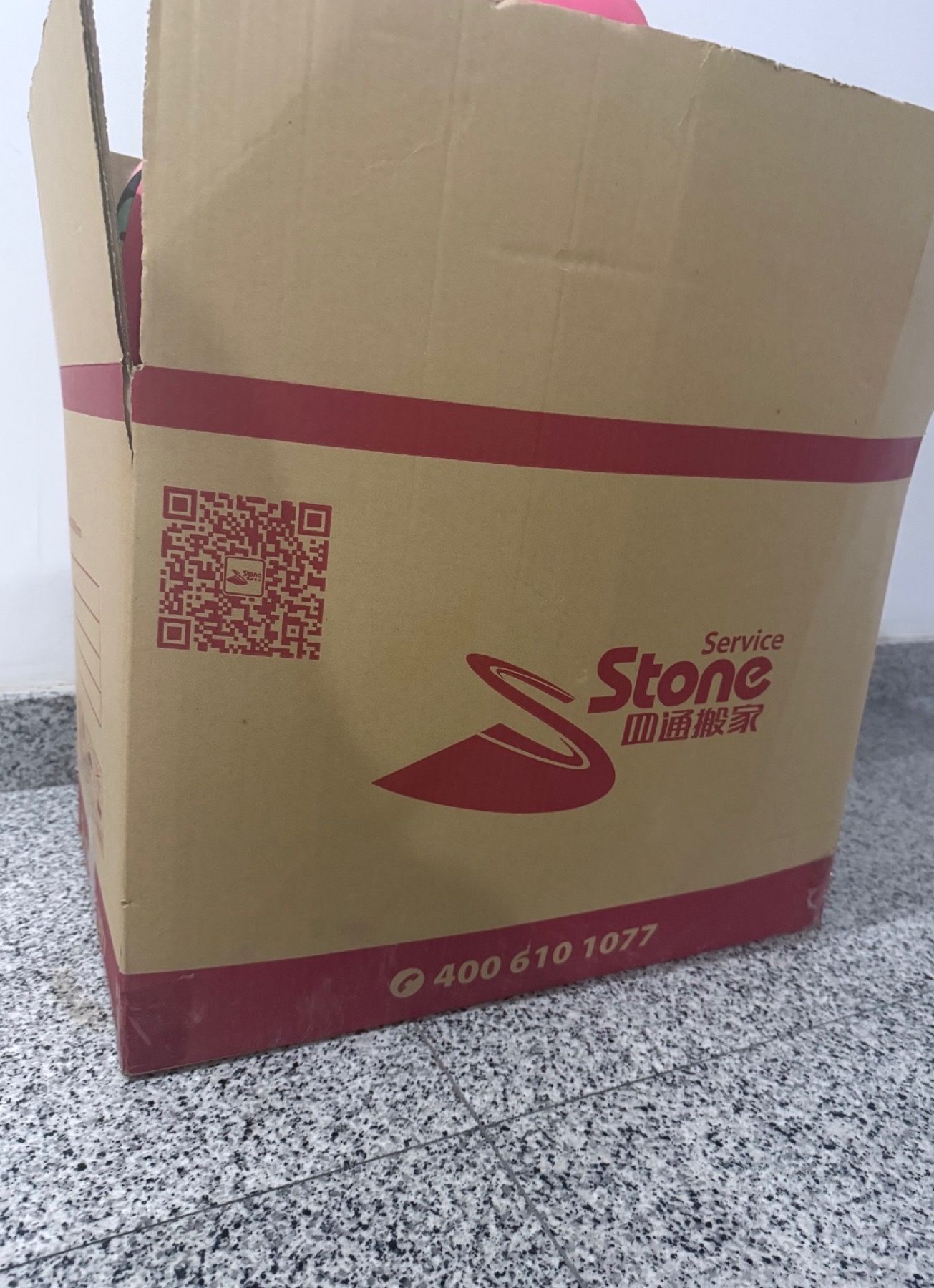}
  \includegraphics[height=0.42\columnwidth]{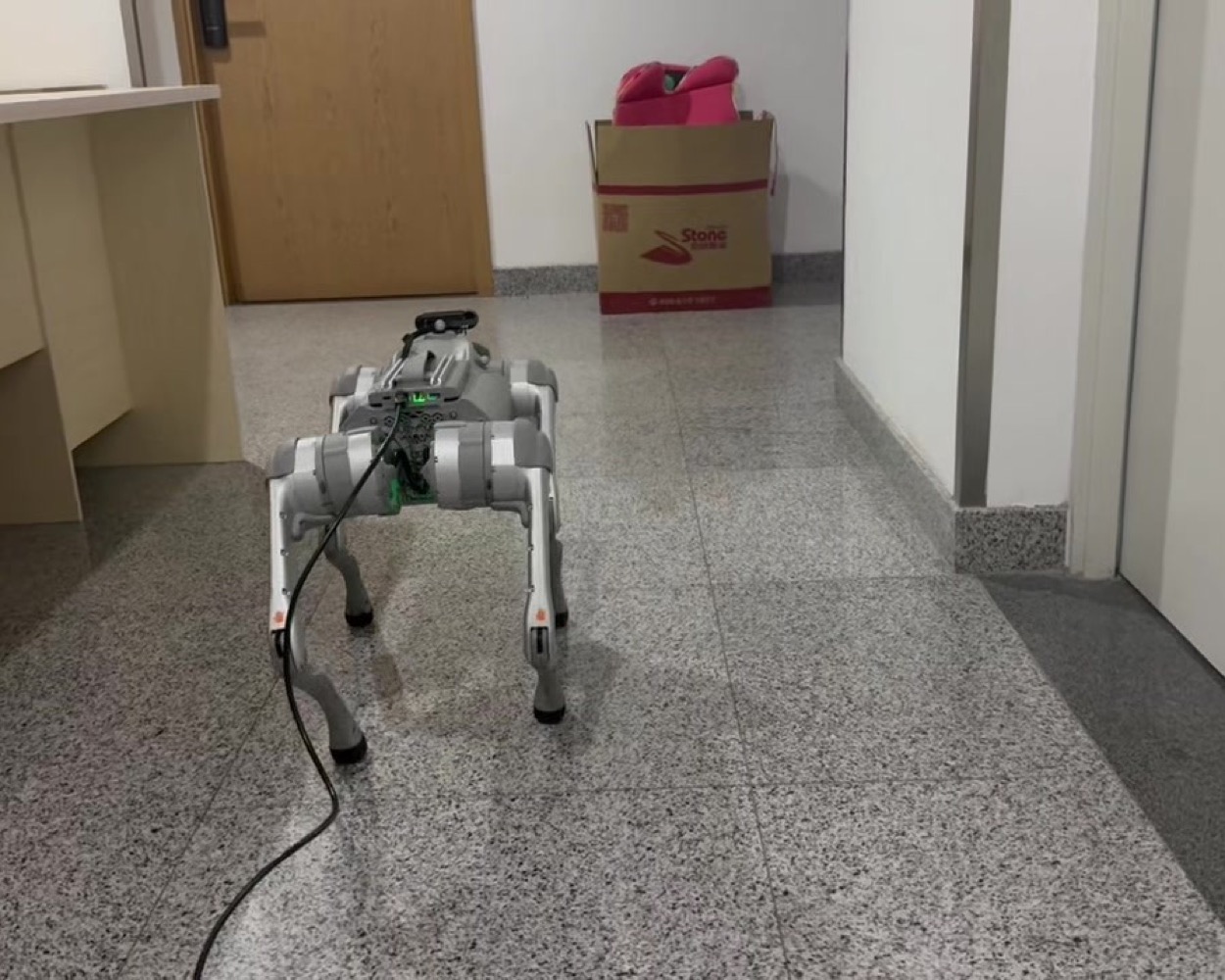}\par
  {\scriptsize (d) Box}\par
  \includegraphics[height=0.42\columnwidth]{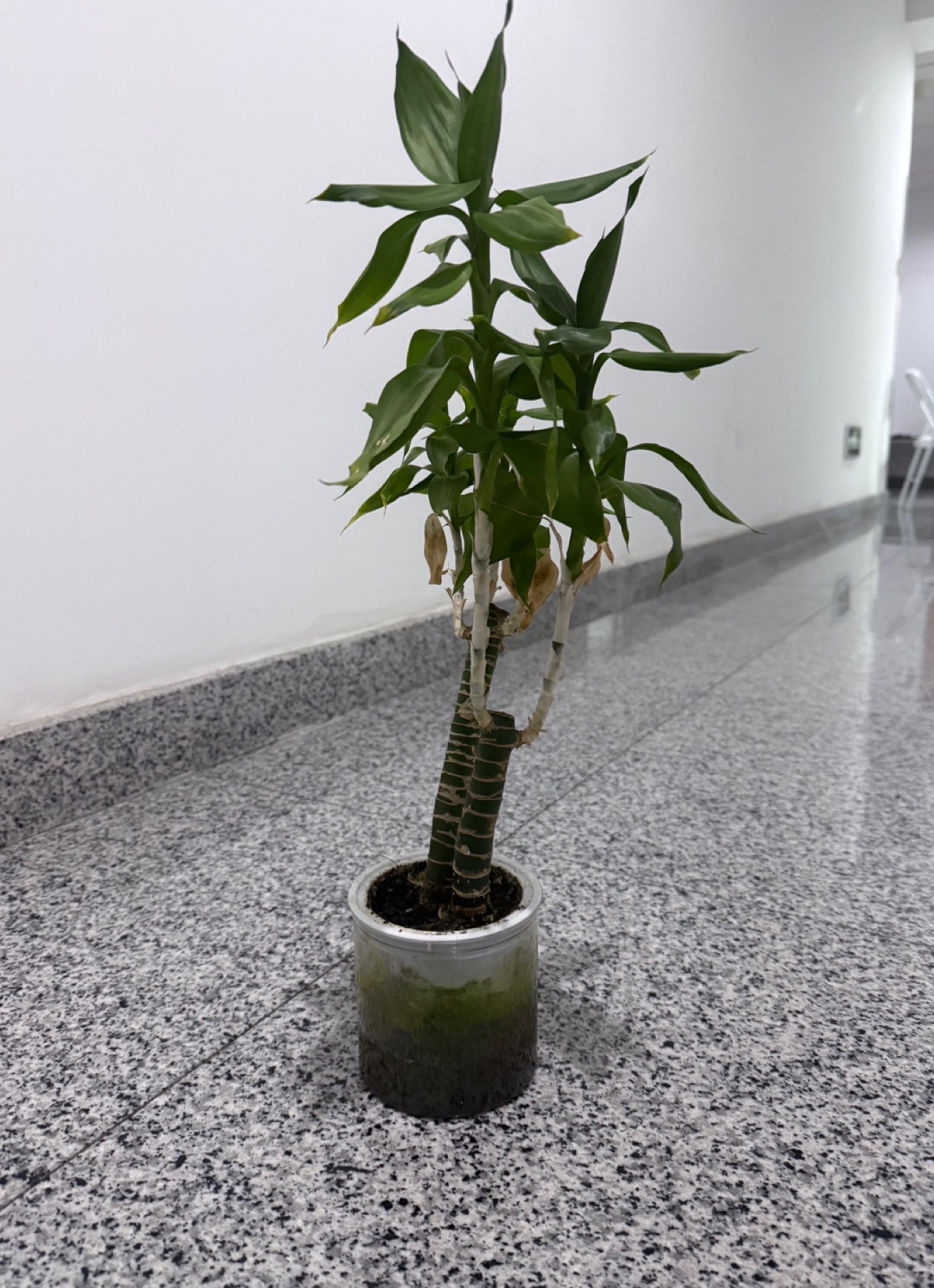}
  \includegraphics[height=0.42\columnwidth]{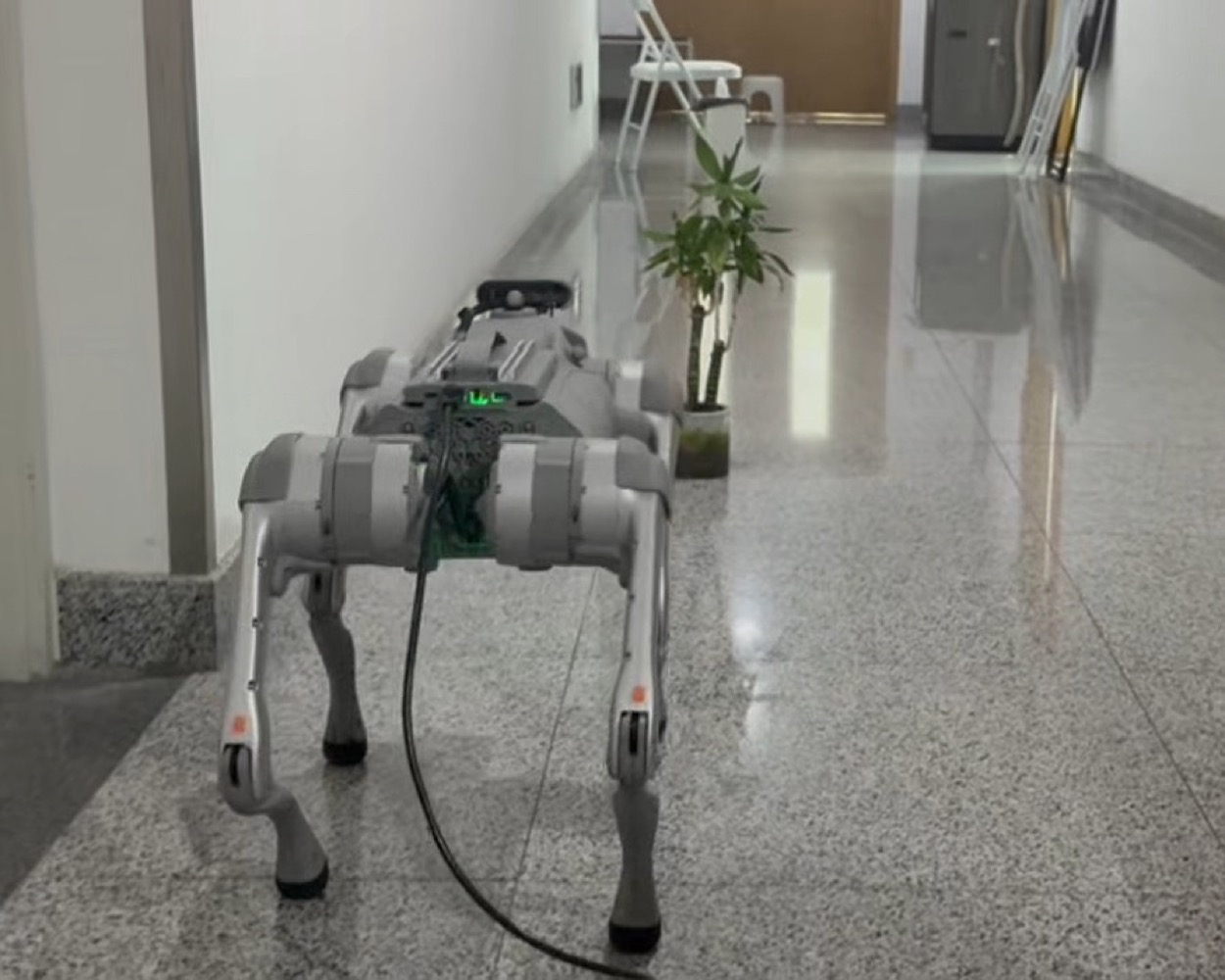}\par
  {\scriptsize (e) Plant}
  \caption{Physical target instances (left) paired with representative
  robot--target views (right): (a) Chair, (b) Table, (c) Trash Bin, (d) Box,
  and (e) Plant.}
  \label{fig:deployment-views}
\end{figure}

As shown in Figure~\ref{fig:deployment-views}, each physical target instance is
paired with a representative robot--target view. The indoor test area combines
reflective hard flooring, glass and metal doors, narrow passages, and movable
furniture. These surfaces and obstacles introduce appearance and local-geometry
variation that is absent from a target-only photograph.

\subsection{Closed-Loop Action Execution}

The physical adapter exposes four executable commands: move forward, turn
left, turn right, and stop. Look-up and look-down actions are not used on the
robot. Forward motion uses a linear velocity of 0.5 m/s, while turns use an
angular-velocity magnitude of 0.35 rad/s. The control loop runs at 10 Hz. A
stop command sets both linear and angular velocities to zero. The Go2 built-in
obstacle-avoidance mode is disabled.

At policy cycle $t$, a camera frame is encoded with the fixed episode goal and
previous-action symbol. The selected physical command is then passed to the
adapter, and the resulting action symbol enters the temporal context for cycle
$t+1$. This preserves a closed sense--infer--act loop while keeping
robot-specific velocity commands outside the learned policy.

\begin{table}[H]
\centering
\small
\renewcommand{\arraystretch}{0.92}
\caption{Deployment-specific configuration.}
\label{tab:deployment-interface}
\begin{tabular}{@{}p{0.31\columnwidth}p{0.63\columnwidth}@{}}
\toprule
Aspect & Supplementary detail \\
\midrule
Robot & Unitree Go2 EDU \\
Camera & External Intel RealSense D435i; $640\times480$ RGB at 15 fps \\
Camera mount & Approximately 0.45 m high; $0^\circ$ pitch \\
Visual encoding & SigLIP \\
Episode reset & WSLA recurrent state reinitialized before every trial \\
Physical actions & Forward, left, right, and stop; no look actions \\
Motion commands & 0.5 m/s forward; 0.35 rad/s turn; stop zeros both velocities \\
Update rate & 10 Hz \\
Built-in avoidance & Disabled \\
Scene factors & Reflective floor; glass/metal doors; narrow passages; movable furniture \\
Visual record & Five paired target and robot--target views \\
\bottomrule
\end{tabular}
\end{table}

\end{document}